\pdfoutput=1
\documentclass[11pt]{article}

\usepackage[final]{acl} %review

\usepackage{times}
\usepackage{latexsym}
\usepackage{booktabs}
\usepackage{float}
\usepackage{tikz,pgfplots}
\usepackage[upright]{fourier}
\usepackage{tkz-kiviat,numprint,fullpage} 
\usepackage{pgfplotstable} 
\usetikzlibrary{arrows}

\pgfplotsset{compat=1.18}
\usepackage[T1]{fontenc}
\usepackage[utf8]{inputenc}
\usepackage{blindtext}
\usepackage{tcolorbox}

\usepackage{microtype}
\usepackage{tabularx}
\usepackage{subcaption}
\usepackage{multicol}

\usepackage{inconsolata}
\usepackage{amsmath}

\usepackage{graphicx}

\usepackage{xspace}
\usepackage{makecell, cellspace, caption}

\newcommand\ColorBox[1]{\textcolor{#1}{\rule{2ex}{2ex}}}

\def\benchmark{\texttt{SLM-Bench}\xspace}
\def\domains{14\xspace}
\def\datasets{23\xspace}
\def\models{15\xspace}
\def\tasks{9\xspace}
\def\metrics{11\xspace}
\def\configs{4\xspace}
\def\co2{CO\textsubscript{2}\xspace}

\title{\texttt{SLM-Bench}: A Comprehensive Benchmark of Small Language Models \\on Environmental Impacts---Extended Version}

\author{
 \textbf{Nghiem Thanh Pham\textsuperscript{1}},
 \textbf{Tung Kieu\textsuperscript{2}},
 \textbf{Duc-Manh Nguyen\textsuperscript{3,6}},
 \textbf{Son Ha Xuan\textsuperscript{4}},\\
 \textbf{Nghia Duong-Trung\textsuperscript{5,6}},
 \textbf{Danh Le-Phuoc\textsuperscript{3}}
\\
 \textsuperscript{1}FPT University, Vietnam,
 \textsuperscript{2}Aalborg University, Denmark,
 \textsuperscript{3}Technische Universität Berlin, Germany,\\
 \textsuperscript{4}RMIT University, Vietnam,
 \textsuperscript{5}German Research Center for Artificial Intelligence, Germany,
 \textsuperscript{6}HiveIntel GmbH, Germany
\\
 \textsuperscript{1}\texttt{nghiemptce160353@fpt.edu.vn},
 \textsuperscript{2}\texttt{tungkvt@cs.aau.dk},
 \textsuperscript{3}\texttt{\{duc.manh.nguyen,danh.lephuoc\}@tu-berlin.de},\\
 \textsuperscript{4}\texttt{ha.son@rmit.edu.vn},
 \textsuperscript{5}\texttt{nghia\_trung.duong@dfki.de}
\\
\\
}

\begin{document}

\maketitle

\setcounter{page}{1}
\begin{abstract}
Small Language Models (SLMs) offer computational efficiency and accessibility, yet a systematic evaluation of their performance and environmental impact remains lacking.
We introduce \benchmark, the first benchmark specifically designed to assess SLMs across multiple dimensions, including accuracy, computational efficiency, and sustainability metrics.
\benchmark evaluates \models SLMs on \tasks NLP tasks using \datasets datasets spanning \domains domains.
The evaluation is conducted on \configs hardware configurations, providing a rigorous comparison of their effectiveness. 
Unlike prior benchmarks, \benchmark quantifies \metrics metrics across correctness, computation, and consumption, enabling a holistic assessment of efficiency trade-offs. 
Our evaluation considers controlled hardware conditions, ensuring fair comparisons across models.
We develop an open-source benchmarking pipeline with standardized evaluation protocols to facilitate reproducibility and further research. 
Our findings highlight the diverse trade-offs among SLMs, where some models excel in accuracy while others achieve superior energy efficiency.
\benchmark sets a new standard for SLM evaluation, bridging the gap between resource efficiency and real-world applicability.
\end{abstract}

\section{Introduction}
Recent advancements in Language Models (LMs) have profoundly influenced a wide range of domains, including finance~\cite{DBLP:conf/eacl/TheumaS24}, healthcare~\cite{DBLP:conf/www/YangZKXHA24}, and manufacturing~\cite{DBLP:journals/corr/abs-2410-21418}.
These models have demonstrated exceptional capabilities in retaining vast amounts of knowledge~\cite{DBLP:conf/acl/ZhengMQWMLFSC23}, solving highly complex tasks~\cite{DBLP:conf/emnlp/BursztynDDB22}, and conducting intricate reasoning processes~\cite{DBLP:conf/acl/YangGKG024} that closely align with human-level intentions.
Their ability to understand context, generate coherent text, and adapt to diverse applications has made them indispensable tools in both academic research and industry practices.

\begin{figure*}[t]
    \centering
    \includegraphics[width=1.0\linewidth]{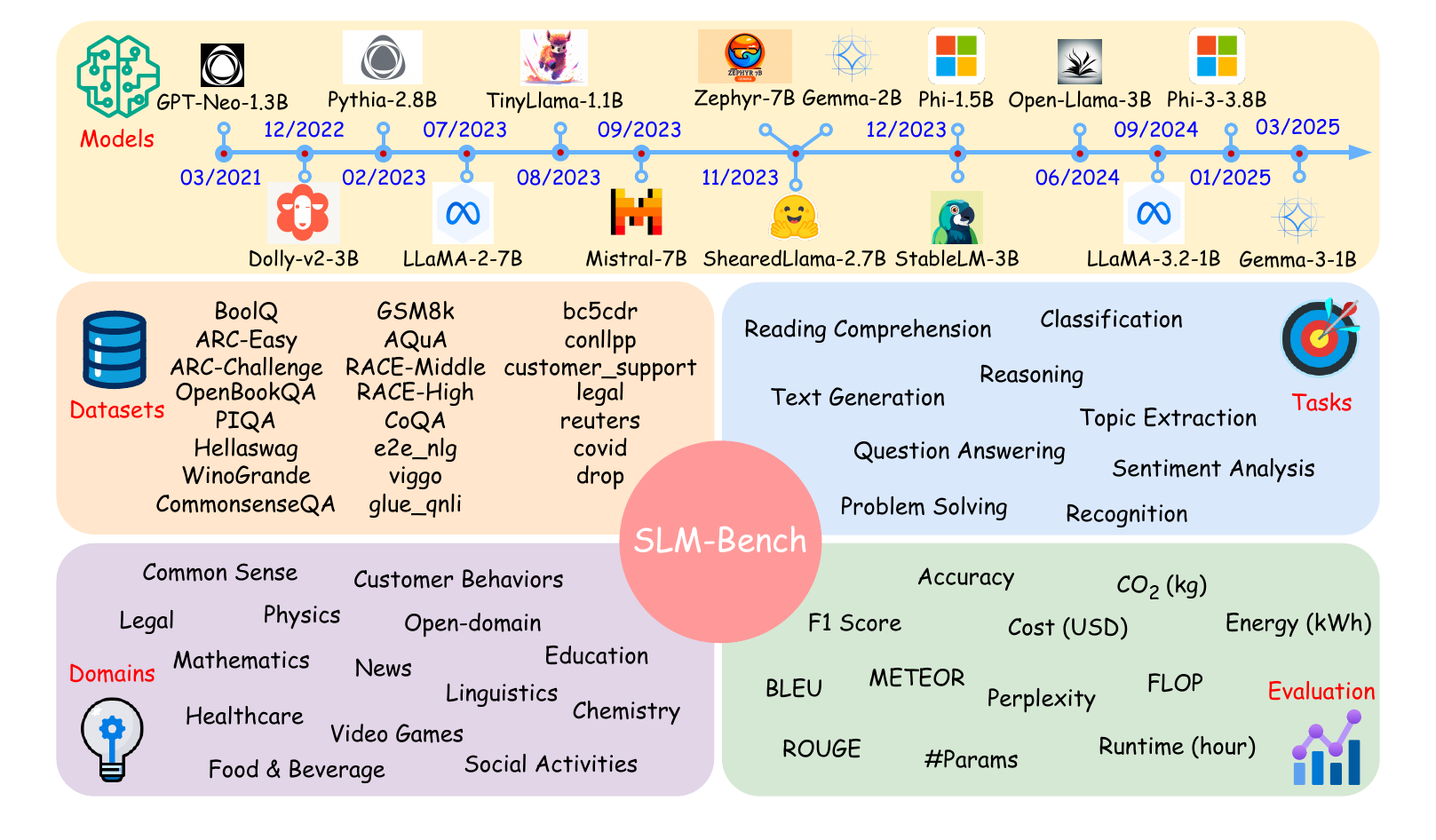}
    \caption{Overview of \benchmark}
    \label{fig:overview}
    % \vspace{-2.0em}
\end{figure*}

However, the impressive performance of LMs often comes at a significant cost.
The large number of parameters that enable their functionality requires extensive computational resources~\cite{DBLP:conf/acl/Lv0LGQ24}, leading to prohibitively high operational expenses.
Additionally, these models demand intensive energy consumption, which not only drives up costs but also contributes to substantial environmental challenges.
The carbon footprint associated with training and deploying Large Language Models (LLMs) has raised concerns about their sustainability~\cite{DBLP:conf/iclr/FaizKWOS0024}, as they emit significant amounts of CO\textsubscript{2}.
These issues underline the pressing need for more efficient and environmentally conscious alternatives to LLMs, particularly in an era of growing awareness around climate change and resource optimization.

SLMs~\cite{DBLP:conf/acl/GaoZLZW23,DBLP:conf/acl/MagisterMAMS23} have emerged as a promising solution to mitigate the negative impacts associated with LLMs.
By significantly reducing the number of parameters, SLMs aim to lower computational costs, minimize energy consumption, and decrease the associated carbon emissions, making them a more sustainable alternative.
Their potential has garnered increasing attention from both academic researchers and industry practitioners, positioning SLMs as a practical choice for applications requiring efficiency and scalability.

Despite their growing prominence, a notable gap exists in the systematic evaluation of SLMs.
Currently, there is no dedicated benchmark that comprehensively assesses the performance of SLMs across diverse tasks while also quantifying their environmental impacts.
This lack of standardized evaluation hinders a deeper understanding of their practical implications, particularly in resource-constrained environments where efficiency and sustainability are paramount.
Addressing this gap is essential to unlock the full potential of SLMs and guide their development and deployment to balance performance with environmental responsibility.

In this paper, we aim to address the existing gap by introducing \benchmark (\textbf{S}mall \textbf{L}anguage \textbf{M}odel-\textbf{Bench}mark), a comprehensive benchmarking framework designed to evaluate SLMs across diverse settings with a focus on their environmental impacts.
The insights gained from \benchmark will be instrumental in guiding the development and deployment of SLM-powered applications, particularly in resource-constrained environments.
\benchmark is characterized by three primary features:
(i) \textbf{Focus on SLMs:} Unlike existing benchmarks that predominantly emphasize LLMs~\cite{DBLP:journals/corr/abs-2406-07545}, \benchmark evaluates SLMs that are computationally efficient and accessible to a wider range of users and systems.
(ii) \textbf{Measurement of Environmental Impacts:} A unique aspect of \benchmark is its integration of metrics for energy consumption and CO\textsubscript{2} emissions.
This allows for a holistic assessment of the sustainability of SLMs, addressing a crucial aspect often overlooked in benchmarking efforts.
(iii) \textbf{Evaluation Across Diverse Settings:} \benchmark rigorously evaluates \models SLMs on \tasks tasks using \datasets datasets from \domains domains.
The evaluation is conducted on \configs hardware configurations, providing a comprehensive analysis of their performance across varied scenarios.
This extensive evaluation ensures a deeper understanding of SLM capabilities and trade-offs.

\benchmark is illustrated in Figure~\ref{fig:overview}, where we provide an overview of its key components, including the selected SLMs, the datasets used, the selected domains, the tasks evaluated, and the metrics employed for assessment.
To the best of our knowledge, \benchmark represents the first systematic benchmarking framework dedicated to the evaluation of SLMs with a specific focus on their environmental impacts.
We hope that \benchmark will drive construction through evaluation, facilitating the development of SLMs in many more applications.
Aiming at reproducibility and transparency, the source code and homepage for \benchmark are provided at \url{https://github.com/slm-bench/slm-bench-experiments} and \url{https://slm-bench.github.io/leaderboard}, respectively which will be continuously updated to catch up the state-the-arts.
In summary, our contributions are as follows.

\begin{itemize}
    \item We introduce \benchmark, a comprehensive benchmark of SLMs, which are computationally efficient and more accessible for deployment in resource-constrained environments.
    \item We consider the environmental impacts, such as energy consumption and CO\textsubscript{2} emissions, enabling a holistic assessment of the sustainability of SLMs.
    \item We extensively evaluate \models SLMs across \tasks tasks using \datasets datasets from \domains domains on \configs hardware configurations, offering a thorough understanding of their capabilities and limitations in various contexts.
\end{itemize}

\section{Related Work}
\label{sec:related-work}

\subsection{Small Language Models}
LLMs, such as \texttt{GPT-4}~\cite{DBLP:journals/corr/abs-2303-08774}, \texttt{LLaMA}~\cite{DBLP:journals/corr/abs-2302-13971}, and \texttt{PaLM}~\cite{DBLP:conf/acl/VilarFCLRF23}, have demonstrated exceptional capabilities across various tasks.
However, their large parameter sizes lead to high computational costs~\cite{DBLP:journals/corr/abs-2004-08900}, energy demands~\cite{DBLP:conf/aaai/StrubellGM20}, and environmental impacts~\cite{DBLP:journals/natmi/Dhar20}.
To address these challenges, SLMs have gained attention for their efficiency and scalability.
Techniques like model pruning~\cite{DBLP:conf/iclr/ZhangBL0HC24,DBLP:conf/iclr/Sun0BK24,DBLP:conf/nips/MaFW23}, knowledge distillation~\cite{DBLP:conf/iclr/Gu0WH24,DBLP:conf/emnlp/ZhangZS0X24}, and low-rank factorization~\cite{DBLP:conf/iclr/HuSWALWWC22,DBLP:conf/iclr/XuXG0CZC0024} have enabled models like \texttt{DistilBERT}~\cite{DBLP:journals/corr/abs-1910-01108} and \texttt{TinyBERT}~\cite{DBLP:conf/emnlp/JiaoYSJCL0L20} to achieve strong performance with fewer parameters.
Existing benchmarks, such as \texttt{GLUE}~\cite{DBLP:conf/iclr/WangSMHLB19} and \texttt{SuperGLUE}~\cite{DBLP:conf/nips/WangPNSMHLB19}, primarily evaluate LLMs and lack a focus on SLMs and their performance.
This leaves a gap in understanding the trade-offs between efficiency, performance, and sustainability. 

\subsection{Emission-aware Benchmarking on Language Models}
\label{subsec:related-work-benchmark}

The development of LMs demands substantial computational resources, raising environmental concerns.
Strubell et al.~\cite{DBLP:conf/acl/StrubellGM19} first quantified LMs' carbon footprint, and later work~\cite{DBLP:conf/aaai/StrubellGM20, DBLP:journals/corr/abs-2104-10350} underscored rising energy demands and sustainability trade-offs.
While tools like \texttt{CodeCarbon}~\cite{DBLP:journals/corr/abs-1911-08354} and \texttt{ML CO\textsubscript{2} Impact}\cite{DBLP:journals/corr/abs-1910-09700} enable emissions tracking, benchmarks such as \texttt{Big-Bench}~\cite{DBLP:journals/tmlr/SrivastavaRRSAF23} focus largely on performance, neglecting energy impact.
Recent work has addressed full-lifecycle emissions.
For instance, Luccioni et al.\cite{DBLP:journals/jmlr/LuccioniVL23} showed that \texttt{BLOOM}'s total emissions (50.5 tCO\textsubscript{2}) exceeded training emissions due to hardware and idle energy.
Hardware choice matters too--T4 to A100 GPU migration can cut emissions by 83\%\cite{DBLP:journals/dai/LiuY24}.
Fine-tuning also incurs significant cost~\cite{DBLP:conf/emnlp/WangNSFL23}.
Gowda et al.~\cite{DBLP:journals/corr/abs-2310-06522} proposed the Sustainable-Accuracy Metric to balance performance and efficiency.
Poddar et al.\cite{DBLP:journals/corr/abs-2502-05610} benchmark LLM inference energy, while Singh et al.\cite{DBLP:journals/corr/abs-2412-04782} explore sustainable training and deployment.
However, these efforts mainly target LLMs and partial aspects of environmental impact.
To our knowledge, \benchmark is the first benchmark focused specifically on SLMs with a comprehensive evaluation of both computational and environmental metrics.

\section{Benchmarking Design}

\subsection{Data Collection}

We extensively collect 23 datasets from 11 diverse domains, including common sense, mathematics, physics, news, and legal, among others.
These datasets encompass 9 task types, covering reading comprehension, text classification, logical reasoning, sentiment analysis, and more.
In total, our dataset collection comprises 799,594 samples, ensuring a well-rounded evaluation framework.
Table~\ref{tab:datasets} provides an overview of the collected datasets along with their associated characteristics.
Due to space limitations, we provide the details and examples of these datasets in Appendices~\ref{app:dataset} and~\ref{app:examples}, respectively.
Our selection of these datasets is motivated by their widespread adoption in previous studies~\cite{DBLP:journals/corr/abs-2406-07545}, ensuring compatibility and comparability with existing benchmarks.
Furthermore, we aim to evaluate SLMs from multiple perspectives, assessing their generalization ability, reasoning capabilities, and robustness across diverse tasks and domains.
Integrating datasets from various fields ensures a comprehensive and challenging benchmark that reflects real-world applications.

\begin{table}[ht!]
    \scriptsize
    \setlength{\tabcolsep}{1.5pt}
    \rowcolors{2}{blue!7}{white}
    \centering
    \resizebox{\columnwidth}{!}{%
    \begin{tabular}{llll}
    \toprule
        \rowcolor{white}
        \textbf{Dataset} & \textbf{\#Samples} & \textbf{Domain} & \textbf{Task} \\
        \midrule
         \textbf{BoolQ} & 15,432 & Open-domain & Question Answering \\
         \textbf{ARC-Easy} & 5,876 & Open-domain & Question Answering \\
         \textbf{ARC-Challenge} & 2,590 & Open-domain & Question Answering \\
         \textbf{OpenBookQA} & 5,957 & Open-domain & Question Answering \\
         \textbf{PIQA} & 16,113 & Physics & Reasoning \\
         \textbf{Hellaswag} & 10,421 & Common Sense & Reasoning \\
         \textbf{WinoGrande} & 44,321 & Common Sense & Reasoning \\
         \textbf{CommonsenseQA} & 12,102 & Common Sense & Reasoning \\
         \textbf{GSM8k} & 8,034 & Mathematics & Problem Solving \\
         \textbf{AQuA} & 99,765 & Mathematics & Problem Solving \\
         \textbf{RACE-Middle} & 24,798 & Education & Reading Comprehension \\
         \textbf{RACE-High} & 26,982 & Education & Reading Comprehension \\
         \textbf{CoQA} & 127,542 & Open-domain & Question Answering \\
         \textbf{e2e\_nlg} & 50,321 & Food \& Beverage & Text Generation \\
         \textbf{viggo} & 9,842 & Video Games & Text Generation \\
         \textbf{glue\_qnli} & 104,543 & Linguistics & Question Answering \\
         \textbf{bc5cdr} & 20,764 & Chemistry & Recognition \\
         \textbf{conllpp} & 23,499 & Linguistics & Recognition \\
         \textbf{customer\_support} & 14,872 & Customer Behaviors & Classification \\
         \textbf{legal} & 49,756 & Legal & Classification \\
         \textbf{reuters} & 9,623 & News & Topic Extraction \\
         \textbf{covid} & 19,874 & Healthcare & Sentiment Analysis \\
         \textbf{drop} & 96,567 & Open-domain & Reasoning \\
        \bottomrule
    \end{tabular}
    }
    \caption{Datasets' Details}
    \label{tab:datasets}
    % \vspace{-2.0em}
\end{table}

\subsection{Model Selection}

We select \models SLMs for our benchmarking based on three key criteria as follows.
(i) \textbf{Model Size}: The primary criterion for classification as an SLM is the number of parameters.
The LMs must have fewer than 7 billion parameters to qualify as SLMs.
(ii) \textbf{Popularity \& Reputation}: The selected SLMs should be widely recognized and developed by well-known organizations to ensure the impact.
(iii) \textbf{Open-Source Availability}: The models must be open-source, allowing for transparency, reproducibility, and further research.
These criteria ensure a fair, representative, and reproducible benchmarking process, focusing on models that are both accessible and widely used in the research community.
Table~\ref{tab:models} provides an overview of the selected models.
Due to space limitations, we provide the details of these models in Appendix~\ref{app:models}.

\begin{table}[ht!]
	\scriptsize
	\setlength{\tabcolsep}{2.8pt}
	\rowcolors{2}{blue!7}{white}
	\centering
	\begin{tabular}{lllcl}
		\toprule
		\rowcolor{white}
		\textbf{Model}            & \textbf{\#Params} (B) & \textbf{Size} (GB) & \textbf{Year} & \textbf{Provider} \\
		\midrule
		\texttt{GPT-Neo-1.3B}     & 1.37                  & 2.46               & 03/2021       & EleutherAI        \\
		\texttt{Dolly-v2-3B}      & 3                     & 5.8                & 12/2022       & Databricks        \\
		\texttt{Pythia-2.8B}      & 2.8                   & 5.5                & 02/2023       & EleutherAI        \\
		\texttt{LLaMA-2-7B}       & 6.47                  & 13                 & 07/2023       & Meta              \\
		\texttt{TinyLlama-1.1B}   & 1.1                   & 2                  & 08/2023       & SUTD              \\
		\texttt{Mistral-7B}       & 7                     & 13                 & 09/2023       & Mistral AI        \\
		\texttt{Zephyr-7B}        & 7                     & 13.74              & 11/2023       & WebPilot.AI       \\
		\texttt{ShearedLlama-2.7B} & 2.7                   & 5                  & 11/2023       & Princeton NLP     \\
		\texttt{Gemma-2B}         & 2                     & 4.67               & 11/2023       & Google            \\
		\texttt{Phi-1.5B}         & 1.42                  & 2.84               & 12/2023       & Microsoft         \\
		\texttt{StableLM-3B}      & 3                     & 6.5                & 12/2023       & Stability AI      \\
		\texttt{Open-LLaMA-3B}    & 3                     & 6.8                & 06/2024       & OpenLM            \\
		\texttt{Llama-3.2-1B}         & 1.24                     & 2.47               & 09/2024    & Meta            \\
		\texttt{Phi-3-3.8B}         & 3.82                     & 2.2               & 01/2025       & Microsoft            \\
		\texttt{Gemma-3-1B}      & 1                   & 2                & 03/2025      & Google        \\
		\bottomrule
	\end{tabular}
	\caption{SLMs' Details (sort by release time)}
	\label{tab:models}
	% \vspace{-2.0em}
\end{table}

\subsection{Evaluation}
\label{subsec:evaluation}
We evaluate an SLM using \metrics metrics, covering various aspects of performance and efficiency.
These metrics assess accuracy, such as \textit{BLEU}, and computational performance, such as \textit{Runtime}.
Our evaluation focuses solely on the fine-tuning process.
Since we do not have access to the necessary data for full training, we rely exclusively on pre-trained models.
For inference, we did not include runtime comparisons in the paper, as the differences between models were insignificant.
Additionally, to measure the resource consumption of SLMs, we incorporate three key metrics: \textit{Cost}, \textit{CO\textsubscript{2}} emissions, and \textit{Energy} usage.
This comprehensive evaluation ensures a balanced assessment of both the effectiveness of the model and its environmental and financial impact.
To measure resource consumption, we use the APIs of the experimental server to measure both FLOPs and cost.
Specifically, we conducted experiments by renting a \textbf{lightning.ai}\footnote{\url{https://lightning.ai/}} server and recorded the deducted amount from our account balance as the cost.
Additionally, we used a built-in function of the same platform to measure FLOPs.
To estimate CO\textsubscript{2} emissions, we use the \texttt{ML CO2} package\footnote{\url{https://mlco2.github.io/impact/}}, calling its built-in function.
For energy consumption, we use the \texttt{Zeus} package\footnote{\url{https://ml.energy/zeus/}} in a similar manner.
It is important to note that FLOP, costs, CO\textsubscript{2} emissions, and energy consumption vary across different hardware configurations.
If end-users run the models on a local server, FLOPs can be measured using the \texttt{calflops} library.
The cost can be estimated by multiplying energy consumption, runtime, and electricity price.
Table~\ref{tab:metrics} provides an overview of the evaluation metrics.
Due to space limitations, we provide the details of these metrics in Appendix~\ref{app:metrics}.

\begin{table}[ht!]
    \scriptsize
    \setlength{\tabcolsep}{1.5pt}
    \rowcolors{2}{blue!7}{white}
    \centering
    \begin{tabular}{p{1.5cm}p{1.5cm}p{2.5cm}p{1.5cm}}
        \toprule
        \rowcolor{white}
        \textbf{Metrics} & \textbf{Evaluation} & \textbf{Task} & \textbf{Dataset} \\
        \midrule
        \textit{Accuracy} & Correctness & Question Answering\newline Classification\newline Recognition\newline Reasoning\newline Problem Solving\newline Reading Comprehension & All\newline except [e2e\_nlg, viggo, reuters] \\
        \textit{F1 Score} & Correctness & Question Answering\newline Classification\newline Recognition\newline Reasoning\newline Problem Solving\newline Reading Comprehension & All\newline except [e2e\_nlg, viggo, reuters] \\
        \textit{BLEU} & Correctness & Text Generation\newline Topic Extraction & e2e\_nlg, viggo, reuters \\
        \textit{ROUGE} & Correctness & Text Generation\newline Topic Extraction & e2e\_nlg, viggo, reuters \\
        \textit{METEOR} & Correctness & Text Generation\newline Topic Extraction & e2e\_nlg, viggo, reuters \\
        \textit{Perplexity} & Correctness & Text Generation\newline Topic Extraction & e2e\_nlg, viggo, reuters \\
        \textit{Runtime} & Computation & All & All \\
        \textit{FLOP} & Computation & All & All \\
        \textit{Cost} & Consumption & All & All \\
        \textit{CO\textsubscript{2}} & Consumption & All & All \\
        \textit{Energy} & Consumption & All & All \\
        \bottomrule
    \end{tabular}
    \caption{Metrics' Details}
    \label{tab:metrics}
    % \vspace{-2.0em}
\end{table}

\subsection{Benchmarking Pipeline}

\begin{figure}[h]
    \centering
    \includegraphics[width=0.9\linewidth]{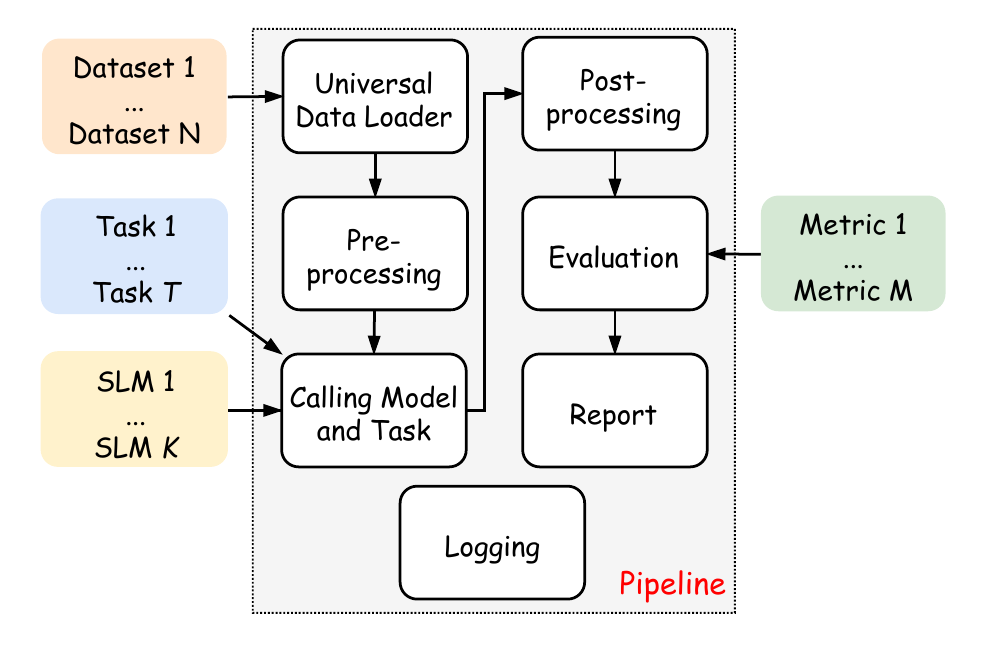}
    \caption{Benchmarking Pipeline}
    \label{fig:pipeline}
    % \vspace{-2.0em}
\end{figure}

To enhance extensibility and reproducibility, we design a unified process pipeline for benchmarking, consisting of 7 key modules.
The \textit{Universal Data Loader} ensures consistency by converting datasets of different formats into a unified structure.
The \textit{Preprocessing} module then refines the data by trimming, removing special symbols, and applying necessary transformations.
Once the data is prepared, the \textit{Calling} module manages the execution of SLMs and tasks, enabling flexibility where a single SLM can be tested on multiple tasks and vice versa.
After inference, the output undergoes further refinement through the \textit{Post-processing} module before moving to the \textit{Evaluation} module, which applies appropriate metrics to assess model performance.
Finally, the \textit{Report} module compiles results and visualizations, providing insights into the model’s effectiveness.
Additionally, the \textit{Logging} module is integrated throughout the pipeline to record all events, enabling better traceability, debugging, and transparency.
This modular design allows for scalability, flexibility, and ease of adaptation, making it well-suited for benchmarking and future extensions.
Figure~\ref{fig:pipeline} illustrates the proposed pipeline.

\subsection{Ranking Methodology}

We propose a ranking method for SLMs based on their performance across multiple datasets and evaluation metrics.
This method counts how often each model ranks first, second, or third in different experimental settings.
Specifically, let $K$ be the number of SLMs, $N$ the number of datasets, $T$ the number of tasks, and $M$ the number of evaluation metrics.
For each model $k$, we count the number of times it achieves the best (gold medal), second-highest (silver medal), and third-highest (bronze medal) performance across all $K \times N \times T \times M$ cases.
This approach provides a straightforward yet effective way to assess overall model performance by considering rankings across diverse conditions rather than relying on a single aggregated metric.
By aggregating the detailed benchmark results into a comprehensive summary, we aim to make insights more accessible to end-users, allowing users to quickly select models based on three major criteria: accuracy, efficiency, or energy consumption.

\section{Experiments}
\label{sec:experiments}

\subsection{Implementation Details}
\label{subsec:implementation-details}

We conduct our experiments on \url{lightning.ai}, a cloud-based platform designed as a development studio for building, training, and deploying AI models.
Lightning AI offers support for various hardware configurations, allowing flexibility in experimentation.
We conduct evaluations across four hardware configurations: two server-grade and two edge-device setups.
The server configurations use NVIDIA L4 and A10 GPUs, while the edge-device configurations use NVIDIA Jetson Orin AGX with 16GB and 64GB memory, respectively.
Due to space constraints, the main paper reports results only on the NVIDIA L4 GPU.
Results for the other configurations are provided on the Leaderboad.
Furthermore, while the magnitudes of the results vary on different hardware configurations, the relative ranking among the SLMs remains consistent.
In addition, we implement the benchmarking pipeline (see Figure~\ref{fig:pipeline}) by using \texttt{Python 3.10}, \texttt{Numpy 1.24}, \texttt{Pandas 1.5}, \texttt{PyTorch 2.0}, \texttt{Sklearn 1.2} and \texttt{Zeus-ML 0.7.0}.

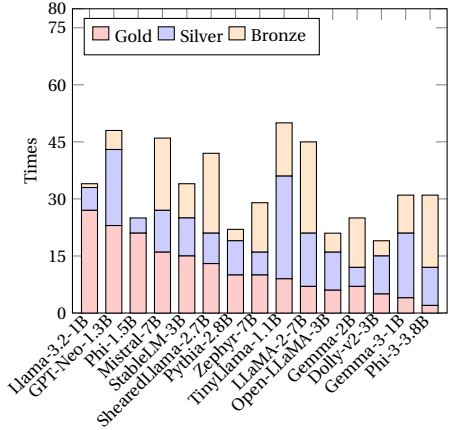
\begin{figure}[t]
    \centering
    \scriptsize
    \begin{tikzpicture}
        \begin{axis}[
            ybar stacked,
            ytick={0, 15, 30, 45, 60, 75, 80},
            yticklabels={0, 15, 30, 45, 60, 75, 80},
            symbolic x coords={Llama-3.2-1B, GPT-Neo-1.3B, Phi-1.5B, Mistral-7B, StableLM-3B, ShearedLlama-2.7B,  Pythia-2.8B, Zephyr-7B, TinyLlama-1.1B, LLaMA-2-7B, Open-LLaMA-3B, Gemma-2B, Dolly-v2-3B, Gemma-3-1B, Phi-3-3.8B},
            xtick=data,
            ymin=0,
            ymax=80,
            ylabel={Times},
            x tick label style={rotate=45, anchor=east},
            width=0.85\linewidth,
            legend pos=north west,
            enlarge x limits=0.05,
            bar width=6pt,
            legend columns=-1,
        ]
        % Data series (stacked bars)
        \addplot[ybar, fill=red!20] table[col sep=comma, x=Model, y=Gold] {data/overall-results-sorted.csv};
        \addplot[ybar, fill=blue!20] table[col sep=comma, x=Model, y=Silver] {data/overall-results-sorted.csv};
        \addplot[ybar, fill=orange!20] table[col sep=comma, x=Model, y=Bronze] {data/overall-results-sorted.csv};
        
        \legend{Gold, Silver, Bronze}
        
        \end{axis}
    \end{tikzpicture}
\caption{Overall Results}
\label{fig:overall-results}
% \vspace{-2.0em}
\end{figure}

\subsection{Hyperparameter Settings}
\label{subsec:hyperparameter-settings}

We select hyperparameters based on recommendations from various sources, including original research papers, official code repositories, and Hugging Face\footnote{https://huggingface.co/} model hubs, which often provide well-tuned configurations for strong performance.
When specific values are unavailable for a given model-dataset pair, we apply a systematic fine-tuning strategy using a validation set to iteratively optimize performance.
Our hyperparameter tuning process includes the following steps: (1) defining appropriate search ranges and (2) conducting a random search over the defined space.
Specifically, we vary \textit{learning\_rate} from 1e-6 to 1e-4; \textit{batch\_size} among {2, 4, 8, 16, 32, 64, 128}; fine-tuning \textit{epochs} among {2, 4, 6, 8, 10}; \textit{LoRA\_rank} among {2, 4, 6, 8, 10}; and dropout probability among {0.1, 0.2, 0.3, 0.4, 0.5}.
This approach ensures a fair and consistent evaluation across all models and datasets.

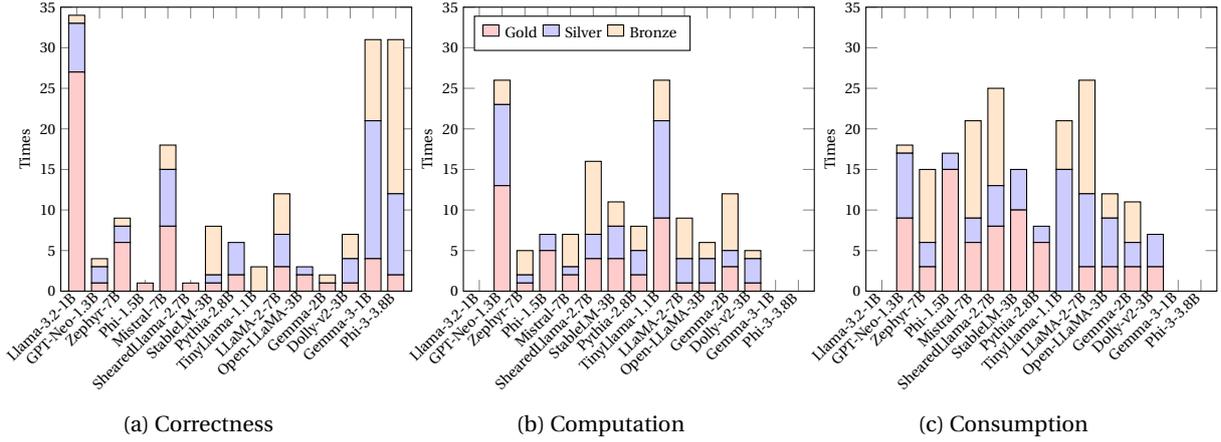
\begin{figure*}[t]
\fontsize{6}{6}\selectfont
\centering
\begin{subfigure}{0.33\linewidth}
    \begin{tikzpicture}
        \begin{axis}[
            ybar stacked,
            ytick={0, 5, 10, 15, 20, 25, 30, 35},
            ytick={0, 5, 10, 15, 20, 25, 30, 35},
            symbolic x coords={Llama-3.2-1B, GPT-Neo-1.3B, Zephyr-7B, Phi-1.5B, Mistral-7B, ShearedLlama-2.7B, StableLM-3B, Pythia-2.8B, TinyLlama-1.1B, LLaMA-2-7B, Open-LLaMA-3B, Gemma-2B, Dolly-v2-3B, Gemma-3-1B, Phi-3-3.8B},
            xtick=data,
            ymin=0,
            ymax=35,
            ylabel={Times},
            x tick label style={rotate=45, anchor=east},
            width=1.18\linewidth,
            legend pos=north east,
            enlarge x limits=0.05,
            bar width=6pt,
            legend columns=-1,
        ]
        % Data series (stacked bars)
        \addplot[ybar, fill=red!20] table[col sep=comma, x=Model, y=Gold] {data/correctness.csv};
        \addplot[ybar, fill=blue!20] table[col sep=comma, x=Model, y=Silver] {data/correctness.csv};
        \addplot[ybar, fill=orange!20] table[col sep=comma, x=Model, y=Bronze] {data/correctness.csv};
        
        % \legend{Gold, Silver, Bronze}
        
        \end{axis}
    \end{tikzpicture}
    \caption{Correctness}
    \label{subfig:correctness}
\end{subfigure}
\begin{subfigure}{0.33\linewidth}
    \begin{tikzpicture}
        \begin{axis}[
            ybar stacked,
            ytick={0, 5, 10, 15, 20, 25, 30, 35},
            ytick={0, 5, 10, 15, 20, 25, 30, 35},
            symbolic x coords={Llama-3.2-1B, GPT-Neo-1.3B, Zephyr-7B, Phi-1.5B, Mistral-7B, ShearedLlama-2.7B, StableLM-3B, Pythia-2.8B, TinyLlama-1.1B, LLaMA-2-7B, Open-LLaMA-3B, Gemma-2B, Dolly-v2-3B, Gemma-3-1B, Phi-3-3.8B},
            xtick=data,
            ymin=0,
            ymax=35,
            ylabel={Times},
            x tick label style={rotate=45, anchor=east},
            width=1.18\linewidth,
            % legend style={at={(1,1)}, anchor=north west,
            legend pos=north west,
            enlarge x limits=0.05,
            bar width=6pt,
            legend columns=-1,
        ]
        % Data series (stacked bars)
        \addplot[ybar, fill=red!20] table[col sep=comma, x=Model, y=Gold] {data/computation.csv};
        \addplot[ybar, fill=blue!20] table[col sep=comma, x=Model, y=Silver] {data/computation.csv};
        \addplot[ybar, fill=orange!20] table[col sep=comma, x=Model, y=Bronze] {data/computation.csv};
        
        \legend{Gold, Silver, Bronze}
        
        \end{axis}
    \end{tikzpicture}
    \caption{Computation}
    \label{subfig:computation}
\end{subfigure}
\begin{subfigure}{0.33\linewidth}
    \begin{tikzpicture}
        \begin{axis}[
            ybar stacked,
            ytick={0, 5, 10, 15, 20, 25, 30, 35},
            ytick={0, 5, 10, 15, 20, 25, 30, 35},
            symbolic x coords={Llama-3.2-1B, GPT-Neo-1.3B, Zephyr-7B, Phi-1.5B, Mistral-7B, ShearedLlama-2.7B, StableLM-3B, Pythia-2.8B, TinyLlama-1.1B, LLaMA-2-7B, Open-LLaMA-3B, Gemma-2B, Dolly-v2-3B, Gemma-3-1B, Phi-3-3.8B},
            xtick=data,
            ymin=0,
            ymax=35,
            ylabel={Times},
            x tick label style={rotate=45, anchor=east},
            width=1.18\linewidth,
            legend pos=north west,
            enlarge x limits=0.05,
            bar width=6pt,
            legend columns=-1,
        ]
        % Data series (stacked bars)
        \addplot[ybar, fill=red!20] table[col sep=comma, x=Model, y=Gold] {data/consumption.csv};
        \addplot[ybar, fill=blue!20] table[col sep=comma, x=Model, y=Silver] {data/consumption.csv};
        \addplot[ybar, fill=orange!20] table[col sep=comma, x=Model, y=Bronze] {data/consumption.csv};
        
        % \legend{Gold, Silver, Bronze}
        
        \end{axis}
    \end{tikzpicture}
    \caption{Consumption}
    \label{subfig:consumption}
\end{subfigure}
\caption{Taxonomy Results}
\label{fig:taxonomy-results}
% \vspace{-2.0em}
\end{figure*}

\subsection{Overall Results}

We present the overall results in Figure~\ref{fig:overall-results}, which visualizes the ranking of each SLM across all datasets and evaluation metrics.
The models are sorted from left to right based on the number of gold medals they have achieved.
Our analysis reveals that \texttt{Llama-3.2-1B} outperforms the other SLMs, securing the highest number of gold medals.
This indicates that it consistently delivers top-tier performance across a diverse range of evaluations.
Additionally, \texttt{GPT-Neo-1.3B} emerges as the runner-up in terms of gold medal count, further demonstrating strong performance.
Following closely, \texttt{Phi-1.5B} performs competitively, securing the third-highest number of gold medals.
This model exhibit strong results, often excelling in specific datasets or metrics.
Meanwhile, \texttt{Mistral-7B}, \texttt{TinyLlama-1.1B} and \texttt{LLaMA-2-7B}, while not securing the most gold medals, frequently appear in the top-three rankings.
This suggests that these models maintain a high level of robustness and consistency across various tasks, even if they do not always achieve the top position.

Furthermore, \texttt{TinyLlama-1.1B} and \texttt{LLaMA-2-7B} appear in the middle of the rankings, securing a moderate number of gold medals while frequently earning silver and bronze medals.
Although they do not dominate in terms of gold medal achievements, their balanced performance across different evaluation criteria suggests that they remain competitive across various tasks.
Their ability to accumulate a significant number of medals overall indicates that they can serve as reliable alternatives to the top-performing models, especially in scenarios where consistency across multiple benchmarks is valued.
Meanwhile, \texttt{Dolly-v2-3B}, \texttt{Gemma-3-1B}, and \texttt{Phi-3-3.8B} rank toward the lower end of the leaderboard, achieving the fewest gold medals among the evaluated models.
While their performance is less dominant, they still manage to secure some silver and bronze medals, indicating that they exhibit strengths in certain areas.
These results highlight the varying capabilities of SLMs and the potential trade-offs when selecting a model for specific applications.
Appendix~\ref{app:metric-contributions} further discusses the metric contribution to medals in \models models. 
Further, Appendix~\ref{app:inference-cost} presents the inference cost.

\subsection{Evaluation Taxonomy}
\label{subsec:evaluation-taxonomy}

We categorize the evaluation metrics into three distinct types: correctness, computation, and consumption (see Table~\ref{tab:metrics}).
Figure~\ref{fig:taxonomy-results} presents the performance results for each evaluation type, highlighting the strengths of different SLMs across these dimensions.

In terms of correctness, \texttt{Llama-3.2-1B} earns the most gold medals, making it the most accurate SLM in our evaluation.
\texttt{Mistral-7B} follows closely, with \texttt{Gemma-3-1B} and \texttt{Phi-3-3.8B} also frequently ranking in the top three, reflecting their consistent output quality.

For computational efficiency, \texttt{GPT-Neo-1.3B} leads with the highest number of gold medals, suggesting strong performance in processing speed and resource usage.
\texttt{TinyLlama-1.1B} and \texttt{ShearedLlama-2.7B} also rank highly, demonstrating notable efficiency across tasks.

In terms of resource consumption, \texttt{Phi-1.5B} stands out as the most energy-efficient model.
\texttt{StableLM-3B} and \texttt{GPT-Neo-1.3B} share the second-highest number of gold medals, while \texttt{LLaMA-2.7B} and \texttt{ShearedLlama-2.7B} frequently appear among the top performers, highlighting their sustainable design.

Although computation and energy consumption are often correlated, they are not equivalent.
Some models use more energy to achieve faster runtimes, while others with similar compute may be less efficient due to non-parallelizable operations and architectural bottlenecks. 

We observe that correctness does not strongly correlate with computation or consumption.
This is because accuracy depends not only on model size, but also on the quality and diversity of pre-training data.
Similarly, consumption and computational cost are influenced by more than just model size--factors like model architecture, parallelization efficiency, and computational complexity play a significant role.
For example, \texttt{Llama-3.2-1B}, despite its small size, achieves the highest accuracy but performs poorly in computation and energy consumption--an unexpected yet insightful outcome.

\subsection{Performance Trade-off}

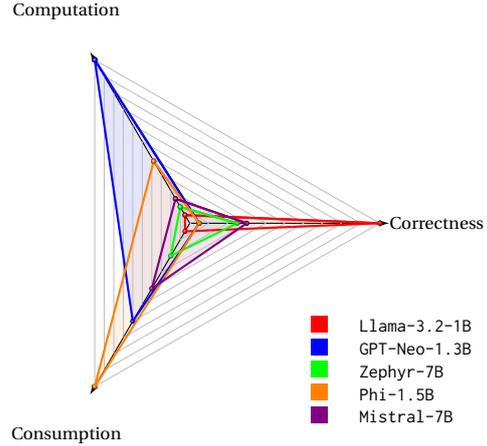
\begin{figure}[t]
    \centering
    \scriptsize
    \begin{tikzpicture}
        \tkzKiviatDiagramFromFile[
                scale=.25,
                label distance=.1cm,
                gap     = 1,label space=3,  
                lattice = 10]{data/radar-results-gold-v2.csv}
        \tkzKiviatLineFromFile[thick,
                               color      = red,
                               mark       = ball,
                               ball color = red,
                               mark size  = 4pt,
                               fill       = red!20]{data/radar-results-gold-v2.csv}{1}
        \tkzKiviatLineFromFile[thick,
                               color      = blue,
                               mark       = ball,
                               ball color = blue,
                               mark size  = 4pt,
                               fill       = blue!20]{data/radar-results-gold-v2.csv}{2}
        \tkzKiviatLineFromFile[thick,
                               color      = green,
                               mark       = ball,
                               ball color = green,
                               mark size  = 4pt,
                               fill       = green!20]{data/radar-results-gold-v2.csv}{3}
        \tkzKiviatLineFromFile[thick,
                               color      = orange,
                               mark       = ball,
                               ball color = orange,
                               mark size  = 4pt,
                               fill       = orange!20]{data/radar-results-gold-v2.csv}{4}
        \tkzKiviatLineFromFile[thick,
                               color      = violet,
                               mark       = ball,
                               ball color = violet,
                               mark size  = 4pt,
                               fill       = violet!20]{data/radar-results-gold-v2.csv}{5}        

        \node[anchor=south west,xshift=-80pt,yshift=5pt] at (current bounding box.south east) 
        {
        \begin{tabular}{@{}lp{3cm}@{}}
        \ColorBox{red} & \texttt{Llama-3.2-1B} \\
        \ColorBox{blue} & \texttt{GPT-Neo-1.3B} \\
        \ColorBox{green} & \texttt{Zephyr-7B} \\
        \ColorBox{orange} & \texttt{Phi-1.5B} \\
        \ColorBox{violet} & \texttt{Mistral-7B} \\
        \end{tabular}
        };
        \end{tikzpicture}
\caption{Performance Trade-off, only Gold Medals}
\label{fig:performance-trade-off-gold}
% \vspace{-1.0em}
\end{figure}

We visualize the performance trade-offs of SLMs using Kiviat charts.
Our evaluation considers three key performance aspects: correctness, computation, and consumption.
Initially, we focus on the number of gold medals each SLM has achieved.
To enhance readability, we select five top-performing SLMs from previous experiments: \texttt{Llama-3.2-1B}, \texttt{GPT-Neo-1.3B}, \texttt{Zephyr-7B}, \texttt{Phi-1.5B}, and \texttt{Mistral-7B}.
Figure~\ref{fig:performance-trade-off-gold} illustrates the performance trade-offs among these models.

Our observations reveal that \texttt{Llama-3.2-1B} excels in correctness but falls short in computation and consumption efficiency.
In contrast, \texttt{Phi-1.5B} demonstrates the highest efficiency in consumption but lags in correctness.
Next, \texttt{GPT-Neo-1.3B} achieves the best performance in computation but also lags in correctness.
\texttt{Mistral-7B} maintains a balanced performance across all three metrics, making it a strong candidate for scenarios requiring an all-around solution.

\begin{figure}[t]
    \centering
    \scriptsize
    \begin{tikzpicture}
        \tkzKiviatDiagramFromFile[
                scale=.25,
                label distance=.1cm,
                gap     = 1,label space=3,  
                lattice = 10]{data/radar-results-score-v2.csv}
        \tkzKiviatLineFromFile[thick,
                               color      = red,
                               mark       = ball,
                               ball color = red,
                               mark size  = 4pt,
                               fill       = red!20]{data/radar-results-score-v2.csv}{1}
        \tkzKiviatLineFromFile[thick,
                               color      = blue,
                               mark       = ball,
                               ball color = blue,
                               mark size  = 4pt,
                               fill       = blue!20]{data/radar-results-score-v2.csv}{2}
        \tkzKiviatLineFromFile[thick,
                               color      = green,
                               mark       = ball,
                               ball color = green,
                               mark size  = 4pt,
                               fill       = green!20]{data/radar-results-score-v2.csv}{3}
        \tkzKiviatLineFromFile[thick,
                               color      = orange,
                               mark       = ball,
                               ball color = orange,
                               mark size  = 4pt,
                               fill       = orange!20]{data/radar-results-score-v2.csv}{4}
        \tkzKiviatLineFromFile[thick,
                               color      = violet,
                               mark       = ball,
                               ball color = violet,
                               mark size  = 4pt,
                               fill       = violet!20]{data/radar-results-score-v2.csv}{5}        

        \node[anchor=south west,xshift=-80pt,yshift=5pt] at (current bounding box.south east) 
        {
        \begin{tabular}{@{}lp{3cm}@{}}
        \ColorBox{red} & \texttt{Llama-3.2-1B} \\
        \ColorBox{blue} & \texttt{GPT-Neo-1.3B} \\
        \ColorBox{green} & \texttt{Zephyr-7B} \\
        \ColorBox{orange} & \texttt{Phi-1.5B} \\
        \ColorBox{violet} & \texttt{Mistral-7B} \\
        \end{tabular}
        };
        \end{tikzpicture}
\caption{Performance Trade-off, Score-based Evaluation}
\label{fig:performance-trade-off-score}
% \vspace{-2.0em}
\end{figure}
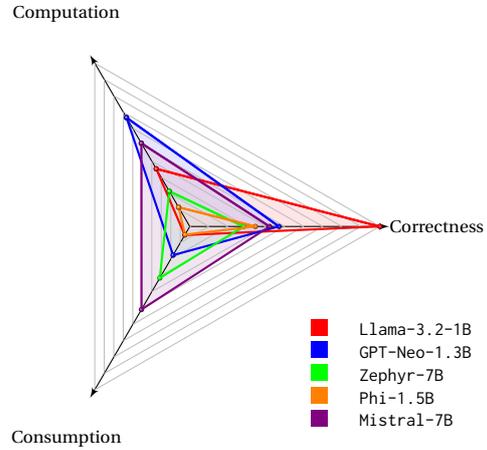

Next, instead of solely considering the number of gold medals, we adopt a score-based evaluation approach that accounts for all gold, silver, and bronze medals.
Specifically, we assign a weighted score to each medal type: each gold medal contributes 3 points, each silver medal contributes 2 points, and each bronze medal contributes 1 point.
We then compute the total score for each SLM across the three evaluation categories by summing the respective scores.
This method provides a more nuanced assessment of model performance, capturing relative strengths beyond just the highest achievements.
To maintain consistency and readability, we again focus on the five top-performing SLMs identified in the previous experiment.
Figure~\ref{fig:performance-trade-off-score} illustrates the performance trade-offs among these models under the score-based evaluation framework.

Our observations remain consistent in that \texttt{Llama-3.2-1B} continues to excel in correctness while exhibiting weaker performance in computation and consumption efficiency.
However, the score-based evaluation reveals additional insights.
Notably, \texttt{Mistral-7B} outperforms \texttt{Phi-1.5B} and \texttt{Zephyr-7B} across all three evaluation metrics, indicating that it provides a more balanced trade-off between accuracy and efficiency.
Furthermore, \texttt{GPT-Neo-1.3B} maintains well-rounded performance across correctness, computation, and consumption, reinforcing its suitability for scenarios where an all-purpose model is desirable.


\subsection{Discussion}

After conducting extensive experiments and analyzing the results, we can derive several key end-user recommendations when selecting an appropriate SLM.
There are clear trade-offs among three key dimensions: correctness, computational efficiency, and resource consumption. 
If accuracy is the primary goal, \texttt{Llama-3.2-1B} is a strong choice, consistently ranking highest in correctness metrics.
However, this comes with increased computational and energy costs. 
For scenarios requiring fast and efficient execution, \texttt{GPT-Neo-1.3B} offers superior runtime performance, making it suitable for latency-sensitive tasks.
If energy efficiency and sustainability are critical, \texttt{Phi-1.5B} is the most resource-friendly option, ideal for deployment on low-power or edge devices. 
For applications needing a balance across all three criteria, \texttt{Mistral-7B} performs reliably and consistently, making it a versatile, well-rounded choice. 
Finally, the choice of SLM should be guided by specific application requirements, as different models offer varying strengths and weaknesses that may influence real-world deployment decisions.

\section{Conclusion}
\label{sec:conclusion}

We introduce \benchmark, a comprehensive benchmark for evaluating the environmental impact of SLMs.
It provides critical insights to guide the selection, optimization, and deployment of models, helping researchers and practitioners balance accuracy, efficiency, and sustainability---especially in resource-constrained settings.
To support future research, \benchmark includes an open-source, extensible pipeline for continuous integration.
In future work, we plan to incorporate additional SLMs and explore a wider range of hardware settings.
Additionally, we aim to enhance the accuracy of energy consumption and CO\textsubscript{2} emission measurements by utilizing sensor-equipped devices.
We also plan to incorporate instruction--following capabilities as a distinct task dimension, extending the benchmark's coverage of real-world use cases.

\section{Limitations}

Although \texttt{SLM-Bench} provides a solid evaluation of SLMs, it has limitations.
First, environmental metrics like energy consumption, CO\textsubscript{2} emissions, and runtime depend heavily on the hardware used.
So far, we have only tested \configs configurations and plan to include more in future work.
Second, energy consumption and CO\textsubscript{2} emissions are currently estimated using standardized formulas rather than real-time measurements.
We aim to improve this by using sensor-equipped devices to capture actual energy usage and emissions.

\section{Broader Impact}

\benchmark promotes the development and deployment of efficient, sustainable, and accessible AI by focusing on SLMs. 
By evaluating models across accuracy, computation, and energy consumption, our benchmark helps users make informed, responsible choices—especially in resource-constrained environments. 
Our inclusion of environmental metrics raises awareness of AI's carbon footprint and supports more sustainable practices. 
By making the benchmark publicly available and regularly updated, we aim to drive progress in green AI and practical model deployment across both academia and industry.

\section{Acknowledgements}
This work was partially funded by the European Union’s programme under grant agreement No.101092908 (SMARTEDGE), by the Chips Joint Undertaking (JU), European Union (EU) HORIZON-JU-IA, under grant agreement No. 101140087 (SMARTY) and by the German Research Foundation (DFG) under the
COSMO project (ref. 453130567).

% \clearpage
% \newpage
\bibliography{custom}

\begin{thebibliography}{60}
\providecommand{\natexlab}[1]{#1}

\bibitem[{Al{-}Shaibani and Ahmad(2023)}]{DBLP:conf/emnlp/Al-ShaibaniA23}
Maged~S. Al{-}Shaibani and Irfan Ahmad. 2023.
\newblock Consonant is all you need: a compact representation of english text for efficient {NLP}.
\newblock In \emph{Findings of the Association for Computational Linguistics (EMNLP)}, pages 11578--11588.

\bibitem[{Authors(2023)}]{DBLP:journals/tmlr/SrivastavaRRSAF23}
Big-Bench Authors. 2023.
\newblock Beyond the imitation game: Quantifying and extrapolating the capabilities of language models.
\newblock \emph{Transaction of Machine Learning Research}, 2023.

\bibitem[{Bisk et~al.(2020)Bisk, Zellers, Bras, Gao, and Choi}]{DBLP:conf/aaai/BiskZLGC20}
Yonatan Bisk, Rowan Zellers, Ronan~Le Bras, Jianfeng Gao, and Yejin Choi. 2020.
\newblock {PIQA:} reasoning about physical commonsense in natural language.
\newblock In \emph{Proceedings of the {AAAI} Conference on Artificial Intelligence (AAAI)}, pages 7432--7439.

\bibitem[{Brabant et~al.(2022)Brabant, Lecorv{\'{e}}, and Rojas{-}Barahona}]{DBLP:conf/lrec/BrabantLB22}
Quentin Brabant, Gw{\'{e}}nol{\'{e}} Lecorv{\'{e}}, and Lina~Maria Rojas{-}Barahona. 2022.
\newblock Coqar: Question rewriting on coqa.
\newblock In \emph{Proceedings of the Language Resources and Evaluation Conference ({LREC})}, pages 119--126.

\bibitem[{Bursztyn et~al.(2022)Bursztyn, Demeter, Downey, and Birnbaum}]{DBLP:conf/emnlp/BursztynDDB22}
Victor~S. Bursztyn, David Demeter, Doug Downey, and Larry Birnbaum. 2022.
\newblock Learning to perform complex tasks through compositional fine-tuning of language models.
\newblock In \emph{Findings of the Association for Computational Linguistics (EMNLP)}, pages 1676--1686.

\bibitem[{Clark et~al.(2019)Clark, Lee, Chang, Kwiatkowski, Collins, and Toutanova}]{DBLP:conf/naacl/ClarkLCK0T19}
Christopher Clark, Kenton Lee, Ming{-}Wei Chang, Tom Kwiatkowski, Michael Collins, and Kristina Toutanova. 2019.
\newblock Boolq: Exploring the surprising difficulty of natural yes/no questions.
\newblock In \emph{Proceedings of the Conference of the North American Chapter of the Association for Computational Linguistics: Human Language Technologies (NAACL-HLT)}, pages 2924--2936.

\bibitem[{Clark et~al.(2018)Clark, Cowhey, Etzioni, Khot, Sabharwal, Schoenick, and Tafjord}]{DBLP:journals/corr/abs-1803-05457}
Peter Clark, Isaac Cowhey, Oren Etzioni, Tushar Khot, Ashish Sabharwal, Carissa Schoenick, and Oyvind Tafjord. 2018.
\newblock Think you have solved question answering? try arc, the {AI2} reasoning challenge.
\newblock \emph{CoRR}, abs/1803.05457.

\bibitem[{Dhar(2020)}]{DBLP:journals/natmi/Dhar20}
Payal Dhar. 2020.
\newblock The carbon impact of artificial intelligence.
\newblock \emph{Nature Machine Intelligence}, 2(8):423--425.

\bibitem[{Dua et~al.(2019)Dua, Wang, Dasigi, Stanovsky, Singh, and Gardner}]{DBLP:conf/naacl/DuaWDSS019}
Dheeru Dua, Yizhong Wang, Pradeep Dasigi, Gabriel Stanovsky, Sameer Singh, and Matt Gardner. 2019.
\newblock {DROP:} {A} reading comprehension benchmark requiring discrete reasoning over paragraphs.
\newblock In \emph{Proceedings of the Conference of the North American Chapter of the Association for Computational Linguistics: Human Language Technologies (NAACL-HLT)}, pages 2368--2378.

\bibitem[{Dusek et~al.(2018)Dusek, Novikova, and Rieser}]{DBLP:conf/inlg/DusekNR18}
Ondrej Dusek, Jekaterina Novikova, and Verena Rieser. 2018.
\newblock Findings of the {E2E} {NLG} challenge.
\newblock In \emph{Proceedings of the International Conference on Natural Language Generation (INLG)}, pages 322--328.

\bibitem[{Faiz et~al.(2024)Faiz, Kaneda, Wang, Osi, Sharma, Chen, and Jiang}]{DBLP:conf/iclr/FaizKWOS0024}
Ahmad Faiz, Sotaro Kaneda, Ruhan Wang, Rita~Chukwunyere Osi, Prateek Sharma, Fan Chen, and Lei Jiang. 2024.
\newblock Llmcarbon: Modeling the end-to-end carbon footprint of large language models.
\newblock In \emph{Proceedings of the International Conference on Learning Representations (ICLR)}.

\bibitem[{Gao et~al.(2023)Gao, Zhou, Liu, Zhao, and Wen}]{DBLP:conf/acl/GaoZLZW23}
Ze{-}Feng Gao, Kun Zhou, Peiyu Liu, Wayne~Xin Zhao, and Ji{-}Rong Wen. 2023.
\newblock Small pre-trained language models can be fine-tuned as large models via over-parameterization.
\newblock In \emph{Proceedings of the Annual Meeting of the Association for Computational Linguistics (ACL)}, pages 3819--3834.

\bibitem[{Gowda et~al.(2023)Gowda, Hao, Li, Sevilla{-}Lara, and Gowda}]{DBLP:journals/corr/abs-2310-06522}
Shreyank~N. Gowda, Xinyue Hao, Gen Li, Laura Sevilla{-}Lara, and Shashank~Narayana Gowda. 2023.
\newblock Watt for what: Rethinking deep learning's energy-performance relationship.
\newblock \emph{CoRR}, abs/2310.06522.

\bibitem[{Gu et~al.(2024)Gu, Dong, Wei, and Huang}]{DBLP:conf/iclr/Gu0WH24}
Yuxian Gu, Li~Dong, Furu Wei, and Minlie Huang. 2024.
\newblock Minillm: Knowledge distillation of large language models.
\newblock In \emph{Proceedings of the International Conference on Learning Representations (ICLR)}.

\bibitem[{Gupta et~al.(2023)Gupta, Agarwal, Mehlawat, Mathur, Somwanshi, and Kumar}]{DBLP:conf/icimmi/GuptaAMMSK23}
Rashmi Gupta, Tushar Agarwal, Aviral Mehlawat, Deeksha Mathur, Devendra~Kumar Somwanshi, and Anil Kumar. 2023.
\newblock {SMER:} {A} novel medical entity recognition technique based on scispacy (bc5cdr) and med7.
\newblock In \emph{Proceedings of the International Conference on Information Management {\&} Machine Intelligence ({ICIMMI})}, pages 119:1--119:6.

\bibitem[{Hu et~al.(2022)Hu, Shen, Wallis, Allen{-}Zhu, Li, Wang, Wang, and Chen}]{DBLP:conf/iclr/HuSWALWWC22}
Edward~J. Hu, Yelong Shen, Phillip Wallis, Zeyuan Allen{-}Zhu, Yuanzhi Li, Shean Wang, Lu~Wang, and Weizhu Chen. 2022.
\newblock Lora: Low-rank adaptation of large language models.
\newblock In \emph{Proceedings of the International Conference on Learning Representations (ICLR)}.

\bibitem[{Jiao et~al.(2020)Jiao, Yin, Shang, Jiang, Chen, Li, Wang, and Liu}]{DBLP:conf/emnlp/JiaoYSJCL0L20}
Xiaoqi Jiao, Yichun Yin, Lifeng Shang, Xin Jiang, Xiao Chen, Linlin Li, Fang Wang, and Qun Liu. 2020.
\newblock Tiny{BERT}: Distilling {BERT} for natural language understanding.
\newblock In \emph{Findings of the Association for Computational Linguistics (EMNLP)}, pages 4163--4174.

\bibitem[{Juraska et~al.(2019)Juraska, Bowden, and Walker}]{DBLP:conf/inlg/JuraskaBW19}
Juraj Juraska, Kevin Bowden, and Marilyn~A. Walker. 2019.
\newblock Viggo: {A} video game corpus for data-to-text generation in open-domain conversation.
\newblock In \emph{Proceedings of the International Conference on Natural Language Generation (INLG)}, pages 164--172.

\bibitem[{Lacoste et~al.(2019)Lacoste, Luccioni, Schmidt, and Dandres}]{DBLP:journals/corr/abs-1910-09700}
Alexandre Lacoste, Alexandra Luccioni, Victor Schmidt, and Thomas Dandres. 2019.
\newblock Quantifying the carbon emissions of machine learning.
\newblock \emph{CoRR}, abs/1910.09700.

\bibitem[{Lai et~al.(2017)Lai, Xie, Liu, Yang, and Hovy}]{DBLP:conf/emnlp/LaiXLYH17}
Guokun Lai, Qizhe Xie, Hanxiao Liu, Yiming Yang, and Eduard~H. Hovy. 2017.
\newblock {RACE:} large-scale reading comprehension dataset from examinations.
\newblock In \emph{Proceedings of the Conference on Empirical Methods in Natural Language Processin (EMNLP)}, pages 785--794.

\bibitem[{Lewis et~al.(2004)Lewis, Yang, Rose, and Li}]{DBLP:journals/jmlr/LewisYRL04}
David~D. Lewis, Yiming Yang, Tony~G. Rose, and Fan Li. 2004.
\newblock {RCV1:} {A} new benchmark collection for text categorization research.
\newblock \emph{Journal of Machine Learning Research}, 5:361--397.

\bibitem[{Li et~al.(2024)Li, Zhao, Jiang, Pan, Liu, Wu, Shu, Tian, Yang, Xu, Lyu, Blenk, Pence, Rupram, Banu, Liu, Wang, Song, Zhai, Song, Zhu, Li, Wang, and Liu}]{DBLP:journals/corr/abs-2410-21418}
Yiwei Li, Huaqin Zhao, Hanqi Jiang, Yi~Pan, Zhengliang Liu, Zihao Wu, Peng Shu, Jie Tian, Tianze Yang, Shaochen Xu, Yanjun Lyu, Parker Blenk, Jacob Pence, Jason Rupram, Eliza Banu, Ninghao Liu, Linbing Wang, Wen{-}Zhan Song, Xiaoming Zhai, Kenan Song, Dajiang Zhu, Beiwen Li, Xianqiao Wang, and Tianming Liu. 2024.
\newblock Large language models for manufacturing.
\newblock \emph{CoRR}, abs/2410.21418.

\bibitem[{Liu et~al.(2023)Liu, Bubeck, Eldan, Kulkarni, Li, Nguyen, Ward, and Zhang}]{DBLP:journals/corr/abs-2312-09241}
Bingbin Liu, S{\'{e}}bastien Bubeck, Ronen Eldan, Janardhan Kulkarni, Yuanzhi Li, Anh Nguyen, Rachel Ward, and Yi~Zhang. 2023.
\newblock Tinygsm: Achieving {\textgreater}80{\%} on gsm8k with small language models.
\newblock \emph{CoRR}, abs/2312.09241.

\bibitem[{Liu and Yin(2024)}]{DBLP:journals/dai/LiuY24}
Vivian Liu and Yiqiao Yin. 2024.
\newblock Green {AI:} exploring carbon footprints, mitigation strategies, and trade-offs in large language model training.
\newblock \emph{Discovery Artificial Intelligence}, 4(1):49.

\bibitem[{Lottick et~al.(2019)Lottick, Susai, Friedler, and Wilson}]{DBLP:journals/corr/abs-1911-08354}
Kadan Lottick, Silvia Susai, Sorelle~A. Friedler, and Jonathan~P. Wilson. 2019.
\newblock Energy usage reports: Environmental awareness as part of algorithmic accountability.
\newblock \emph{CoRR}, abs/1911.08354.

\bibitem[{Luccioni et~al.(2023)Luccioni, Viguier, and Ligozat}]{DBLP:journals/jmlr/LuccioniVL23}
Alexandra~Sasha Luccioni, Sylvain Viguier, and Anne{-}Laure Ligozat. 2023.
\newblock Estimating the carbon footprint of bloom, a 176b parameter language model.
\newblock \emph{Journal of Machine Learning Research}, 24:253:1--253:15.

\bibitem[{Lv et~al.(2024)Lv, Yang, Liu, Guo, and Qiu}]{DBLP:conf/acl/Lv0LGQ24}
Kai Lv, Yuqing Yang, Tengxiao Liu, Qipeng Guo, and Xipeng Qiu. 2024.
\newblock Full parameter fine-tuning for large language models with limited resources.
\newblock In \emph{Proceedings of the Annual Meeting of the Association for Computational Linguistics (ACL)}, pages 8187--8198.

\bibitem[{Ma et~al.(2023)Ma, Fang, and Wang}]{DBLP:conf/nips/MaFW23}
Xinyin Ma, Gongfan Fang, and Xinchao Wang. 2023.
\newblock Llm-pruner: On the structural pruning of large language models.
\newblock In \emph{Proceedings of the Annual Conference on Neural Information Processing Systems ({NeurIPS})}.

\bibitem[{Magister et~al.(2023)Magister, Mallinson, Ad{\'{a}}mek, Malmi, and Severyn}]{DBLP:conf/acl/MagisterMAMS23}
Lucie~Charlotte Magister, Jonathan Mallinson, Jakub Ad{\'{a}}mek, Eric Malmi, and Aliaksei Severyn. 2023.
\newblock Teaching small language models to reason.
\newblock In \emph{Proceedings of the Annual Meeting of the Association for Computational Linguistics (ACL)}, pages 1773--1781.

\bibitem[{Mihaylov et~al.(2018)Mihaylov, Clark, Khot, and Sabharwal}]{DBLP:conf/emnlp/MihaylovCKS18}
Todor Mihaylov, Peter Clark, Tushar Khot, and Ashish Sabharwal. 2018.
\newblock Can a suit of armor conduct electricity? {A} new dataset for open book question answering.
\newblock In \emph{Proceedings of the Conference on Empirical Methods in Natural Language Processing (EMNLP)}, pages 2381--2391.

\bibitem[{Myrzakhan et~al.(2024)Myrzakhan, Bsharat, and Shen}]{DBLP:journals/corr/abs-2406-07545}
Aidar Myrzakhan, Sondos~Mahmoud Bsharat, and Zhiqiang Shen. 2024.
\newblock Open-llm-leaderboard: From multi-choice to open-style questions for llms evaluation, benchmark, and arena.
\newblock \emph{CoRR}, abs/2406.07545.

\bibitem[{Naseem et~al.(2021)Naseem, Razzak, Khushi, Eklund, and Kim}]{DBLP:journals/tcss/NaseemRKEK21}
Usman Naseem, Imran Razzak, Matloob Khushi, Peter~W. Eklund, and Jinman Kim. 2021.
\newblock Covidsenti: {A} large-scale benchmark twitter data set for {COVID-19} sentiment analysis.
\newblock \emph{{IEEE} Transactions on Computational Social Systems}, 8(4):1003--1015.

\bibitem[{OpenAI(2023)}]{DBLP:journals/corr/abs-2303-08774}
OpenAI. 2023.
\newblock {GPT-4} technical report.
\newblock \emph{CoRR}, abs/2303.08774.

\bibitem[{Patterson et~al.(2021)Patterson, Gonzalez, Le, Liang, Munguia, Rothchild, So, Texier, and Dean}]{DBLP:journals/corr/abs-2104-10350}
David~A. Patterson, Joseph Gonzalez, Quoc~V. Le, Chen Liang, Lluis{-}Miquel Munguia, Daniel Rothchild, David~R. So, Maud Texier, and Jeff Dean. 2021.
\newblock Carbon emissions and large neural network training.
\newblock \emph{CoRR}, abs/2104.10350.

\bibitem[{Peng et~al.(2023)Peng, Chen, Shou, Jiang, and Chen}]{DBLP:journals/pvldb/Peng0SJ023}
Yuchen Peng, Ke~Chen, Lidan Shou, Dawei Jiang, and Gang Chen. 2023.
\newblock {AQUA:} automatic collaborative query processing in analytical database.
\newblock \emph{Proc. {VLDB} Endow.}, 16(12):4006--4009.

\bibitem[{Poddar et~al.(2025)Poddar, Koley, Misra, Ganguly, and Ghosh}]{DBLP:journals/corr/abs-2502-05610}
Soham Poddar, Paramita Koley, Janardan Misra, Niloy Ganguly, and Saptarshi Ghosh. 2025.
\newblock Towards sustainable {NLP:} insights from benchmarking inference energy in large language models.
\newblock \emph{CoRR}, abs/2502.05610.

\bibitem[{Prabhu et~al.(2022)Prabhu, Brahma, and Misra}]{DBLP:conf/comad/0002BM22}
Sumanth Prabhu, Aditya~Kiran Brahma, and Hemant Misra. 2022.
\newblock Customer support chat intent classification using weak supervision and data augmentation.
\newblock In \emph{Proceedings of the Joint International Conference on Data Science {\&} Management of Data (CODS-COMAD)}, pages 144--152.

\bibitem[{Sakaguchi et~al.(2020)Sakaguchi, Bras, Bhagavatula, and Choi}]{DBLP:conf/aaai/SakaguchiBBC20}
Keisuke Sakaguchi, Ronan~Le Bras, Chandra Bhagavatula, and Yejin Choi. 2020.
\newblock Winogrande: An adversarial winograd schema challenge at scale.
\newblock In \emph{Proceedings of the {AAAI} Conference on Artificial Intelligence (AAAI)}, pages 8732--8740.

\bibitem[{Sanh et~al.(2019)Sanh, Debut, Chaumond, and Wolf}]{DBLP:journals/corr/abs-1910-01108}
Victor Sanh, Lysandre Debut, Julien Chaumond, and Thomas Wolf. 2019.
\newblock Distil{BERT}, a distilled version of {BERT:} smaller, faster, cheaper and lighter.
\newblock \emph{CoRR}, abs/1910.01108.

\bibitem[{Sharir et~al.(2020)Sharir, Peleg, and Shoham}]{DBLP:journals/corr/abs-2004-08900}
Or~Sharir, Barak Peleg, and Yoav Shoham. 2020.
\newblock The cost of training {NLP} models: {A} concise overview.
\newblock \emph{CoRR}, abs/2004.08900.

\bibitem[{Shu and Du(2024)}]{DBLP:journals/corr/abs-2410-23099}
Dong Shu and Mengnan Du. 2024.
\newblock Comparative analysis of demonstration selection algorithms for {LLM} in-context learning.
\newblock \emph{CoRR}, abs/2410.23099.

\bibitem[{Singh et~al.(2024)Singh, Patel, Ehtesham, Kumar, and Khoei}]{DBLP:journals/corr/abs-2412-04782}
Aditi Singh, Nirmal~Prakashbhai Patel, Abul Ehtesham, Saket Kumar, and Tala~Talaei Khoei. 2024.
\newblock A survey of sustainability in large language models: Applications, economics, and challenges.
\newblock \emph{CoRR}, abs/2412.04782.

\bibitem[{Strubell et~al.(2019)Strubell, Ganesh, and McCallum}]{DBLP:conf/acl/StrubellGM19}
Emma Strubell, Ananya Ganesh, and Andrew McCallum. 2019.
\newblock Energy and policy considerations for deep learning in {NLP}.
\newblock In \emph{Proceedings of the Conference of the Association for Computational Linguistics (ACL)}, pages 3645--3650.

\bibitem[{Strubell et~al.(2020)Strubell, Ganesh, and McCallum}]{DBLP:conf/aaai/StrubellGM20}
Emma Strubell, Ananya Ganesh, and Andrew McCallum. 2020.
\newblock Energy and policy considerations for modern deep learning research.
\newblock In \emph{Proceedings of the {AAAI} Conference on Artificial Intelligence (AAAI)}, pages 13693--13696.

\bibitem[{Sun et~al.(2024)Sun, Liu, Bair, and Kolter}]{DBLP:conf/iclr/Sun0BK24}
Mingjie Sun, Zhuang Liu, Anna Bair, and J.~Zico Kolter. 2024.
\newblock A simple and effective pruning approach for large language models.
\newblock In \emph{Proceedings of the International Conference on Learning Representations (ICLR)}.

\bibitem[{Theuma and Shareghi(2024)}]{DBLP:conf/eacl/TheumaS24}
Adrian Theuma and Ehsan Shareghi. 2024.
\newblock Equipping language models with tool use capability for tabular data analysis in finance.
\newblock In \emph{Proceedings of the Conference of the European Chapter of the Association for Computational Linguistics (EACL)}, pages 90--103.

\bibitem[{Touvron et~al.(2023)Touvron, Lavril, Izacard, Martinet, Lachaux, Lacroix, Rozi{\`{e}}re, Goyal, Hambro, Azhar, Rodriguez, Joulin, Grave, and Lample}]{DBLP:journals/corr/abs-2302-13971}
Hugo Touvron, Thibaut Lavril, Gautier Izacard, Xavier Martinet, Marie{-}Anne Lachaux, Timoth{\'{e}}e Lacroix, Baptiste Rozi{\`{e}}re, Naman Goyal, Eric Hambro, Faisal Azhar, Aur{\'{e}}lien Rodriguez, Armand Joulin, Edouard Grave, and Guillaume Lample. 2023.
\newblock {LLaMA}: Open and efficient foundation language models.
\newblock \emph{CoRR}, abs/2302.13971.

\bibitem[{Vilar et~al.(2023)Vilar, Freitag, Cherry, Luo, Ratnakar, and Foster}]{DBLP:conf/acl/VilarFCLRF23}
David Vilar, Markus Freitag, Colin Cherry, Jiaming Luo, Viresh Ratnakar, and George~F. Foster. 2023.
\newblock Prompting palm for translation: Assessing strategies and performance.
\newblock In \emph{Proceedings of the Annual Meeting of the Association for Computational Linguistics (ACL)}, pages 15406--15427.

\bibitem[{Wan et~al.(2019)Wan, Papageorgiou, Seddon, and Bernardoni}]{DBLP:journals/corr/abs-1912-06905}
Lulu Wan, George Papageorgiou, Michael Seddon, and Mirko Bernardoni. 2019.
\newblock Long-length legal document classification.
\newblock \emph{CoRR}, abs/1912.06905.

\bibitem[{Wang et~al.(2019{\natexlab{a}})Wang, Pruksachatkun, Nangia, Singh, Michael, Hill, Levy, and Bowman}]{DBLP:conf/nips/WangPNSMHLB19}
Alex Wang, Yada Pruksachatkun, Nikita Nangia, Amanpreet Singh, Julian Michael, Felix Hill, Omer Levy, and Samuel~R. Bowman. 2019{\natexlab{a}}.
\newblock {SuperGLUE}: {A} stickier benchmark for general-purpose language understanding systems.
\newblock In \emph{Proceedings of the Annual Conference on Neural Information Processing Systems (NeurIPS)}, pages 3261--3275.

\bibitem[{Wang et~al.(2019{\natexlab{b}})Wang, Singh, Michael, Hill, Levy, and Bowman}]{DBLP:conf/iclr/WangSMHLB19}
Alex Wang, Amanpreet Singh, Julian Michael, Felix Hill, Omer Levy, and Samuel~R. Bowman. 2019{\natexlab{b}}.
\newblock {GLUE}: {A} multi-task benchmark and analysis platform for natural language understanding.
\newblock In \emph{Proceedings of the International Conference on Learning Representations, (ICLR)}.

\bibitem[{Wang et~al.(2023)Wang, Na, Strubell, Friedler, and Luccioni}]{DBLP:conf/emnlp/WangNSFL23}
Xiaorong Wang, Clara Na, Emma Strubell, Sorelle Friedler, and Sasha Luccioni. 2023.
\newblock Energy and carbon considerations of fine-tuning {BERT}.
\newblock In \emph{Findings of the Association for Computational Linguistics ({EMNLP})}, pages 9058--9069.

\bibitem[{Xu et~al.(2024)Xu, Xie, Gu, Chen, Chang, Zhang, Chen, Zhang, and Tian}]{DBLP:conf/iclr/XuXG0CZC0024}
Yuhui Xu, Lingxi Xie, Xiaotao Gu, Xin Chen, Heng Chang, Hengheng Zhang, Zhengsu Chen, Xiaopeng Zhang, and Qi~Tian. 2024.
\newblock Qa-lora: Quantization-aware low-rank adaptation of large language models.
\newblock In \emph{Proceedings of the International Conference on Learning Representations (ICLR)}.

\bibitem[{Yang et~al.(2024{\natexlab{a}})Yang, Zhang, Kuang, Xie, Huang, and Ananiadou}]{DBLP:conf/www/YangZKXHA24}
Kailai Yang, Tianlin Zhang, Ziyan Kuang, Qianqian Xie, Jimin Huang, and Sophia Ananiadou. 2024{\natexlab{a}}.
\newblock {MentaLLaMA}: Interpretable mental health analysis on social media with large language models.
\newblock In \emph{Proceedings of the {ACM} on Web Conference (WWW)}, pages 4489--4500.

\bibitem[{Yang et~al.(2024{\natexlab{b}})Yang, Gribovskaya, Kassner, Geva, and Riedel}]{DBLP:conf/acl/YangGKG024}
Sohee Yang, Elena Gribovskaya, Nora Kassner, Mor Geva, and Sebastian Riedel. 2024{\natexlab{b}}.
\newblock Do large language models latently perform multi-hop reasoning?
\newblock In \emph{Proceedings of the Annual Meeting of the Association for Computational Linguistics (ACL)}, pages 10210--10229.

\bibitem[{Zellers et~al.(2019)Zellers, Holtzman, Bisk, Farhadi, and Choi}]{DBLP:conf/acl/ZellersHBFC19}
Rowan Zellers, Ari Holtzman, Yonatan Bisk, Ali Farhadi, and Yejin Choi. 2019.
\newblock Hellaswag: Can a machine really finish your sentence?
\newblock In \emph{Proceedings of the Conference of the Association for Computational Linguistics (ACL)}, pages 4791--4800.

\bibitem[{Zhang and Li(2023)}]{DBLP:conf/iconip/ZhangL23}
Lihui Zhang and Ruifan Li. 2023.
\newblock Knowledge prompting with contrastive learning for unsupervised commonsenseqa.
\newblock In \emph{Neural Information Processing - Proceedings of the International Conference on Neural Information Processing ({ICONIP})}, pages 27--38.

\bibitem[{Zhang et~al.(2024{\natexlab{a}})Zhang, Zhang, Sun, Chen, and Xu}]{DBLP:conf/emnlp/ZhangZS0X24}
Songming Zhang, Xue Zhang, Zengkui Sun, Yufeng Chen, and Jinan Xu. 2024{\natexlab{a}}.
\newblock Dual-space knowledge distillation for large language models.
\newblock In \emph{Proceedings of the Conference on Empirical Methods in Natural Language Processing (EMNLP)}, pages 18164--18181.

\bibitem[{Zhang et~al.(2024{\natexlab{b}})Zhang, Bai, Lin, Zhao, Hou, and Cannistraci}]{DBLP:conf/iclr/ZhangBL0HC24}
Yingtao Zhang, Haoli Bai, Haokun Lin, Jialin Zhao, Lu~Hou, and Carlo~Vittorio Cannistraci. 2024{\natexlab{b}}.
\newblock Plug-and-play: An efficient post-training pruning method for large language models.
\newblock In \emph{Proceedings of the International Conference on Learning Representations (ICLR)}.

\bibitem[{Zheng et~al.(2023)Zheng, Ma, Qiu, Wu, Ma, Liu, Feng, Shang, and Chen}]{DBLP:conf/acl/ZhengMQWMLFSC23}
Junhao Zheng, Qianli Ma, Shengjie Qiu, Yue Wu, Peitian Ma, Junlong Liu, Huawen Feng, Xichen Shang, and Haibin Chen. 2023.
\newblock Preserving commonsense knowledge from pre-trained language models via causal inference.
\newblock In \emph{Proceedings of the Annual Meeting of the Association for Computational Linguistics (ACL)}, pages 9155--9173.

\end{thebibliography}

\clearpage
\newpage
\setcounter{page}{1}
\appendix
\setcounter{figure}{0}
\renewcommand{\figurename}{Figure}
\renewcommand{\thefigure}{A\arabic{figure}}
\setcounter{table}{0}
\renewcommand{\tablename}{Table}
\renewcommand{\thetable}{A\arabic{table}}

\begin{table*}[t]
    \renewcommand{\arraystretch}{1.2}
    \rowcolors{2}{blue!7}{white}
    \fontsize{7.4}{7.4}\selectfont
    \setlength{\tabcolsep}{1.8pt}
    \centering
    \begin{tabularx}{\textwidth}{lllllllX}
    \toprule
        \rowcolor{white}
        \textbf{No} & \textbf{Dataset} & \textbf{\#Samples} & \textbf{\#Tokens} & \textbf{Domain} & \textbf{Task} & \textbf{Venues} & \textbf{Description} \\
        \midrule
        1 & \textbf{BoolQ} & 15,432 & 3M & Common Sense & Question Answering & NAACL 2019 & A yes/no question-answering dataset where each question requires reasoning over a short passage~\cite{DBLP:conf/naacl/ClarkLCK0T19}.\\
        2 & \textbf{ARC-Easy} & 5,876 & 1.8M & Common Sense & Question Answering & arXiv & A subset of the AI2 Reasoning Challenge with straightforward grade-school science questions~\cite{DBLP:journals/corr/abs-1803-05457}.\\
        3 & \textbf{ARC-Challenge} & 2,590 & 1.5M & Common Sense & Question Answering & arXiv & A more difficult subset of ARC with questions requiring reasoning and external knowledge~\cite{DBLP:journals/corr/abs-1803-05457}.\\
        4 & \textbf{OpenBookQA} & 5,957 & 1M & Common Sense & Question Answering & EMNLP 2018 & A multiple-choice question-answering dataset that requires knowledge from a small ``open book'' of science facts~\cite{DBLP:conf/emnlp/MihaylovCKS18}.\\
        5 & \textbf{PIQA} & 16,113 & 1.6M & Physics & Reasoning & AAAI 2020 & A dataset for physical commonsense reasoning, testing knowledge of how objects are used in everyday situations~\cite{DBLP:conf/aaai/BiskZLGC20}. \\
        6 & \textbf{Hellaswag} & 10,421 & 10M & Common Sense & Reasoning & ACL 2019 & A dataset for commonsense reasoning and next-sentence prediction with adversarially mined, plausible distractors~\cite{DBLP:conf/acl/ZellersHBFC19}. \\
        7 & \textbf{WinoGrande} & 44,321 & 5M & Common Sense & Reasoning & AAAI 2020 & A large-scale dataset for pronoun resolution tasks, inspired by the Winograd Schema Challenge~\cite{DBLP:conf/aaai/SakaguchiBBC20}. \\
        8 & \textbf{CommonsenseQA} & 12,102 & 1.8M & Common Sense & Reasoning & ICONIP 2023 & A multiple-choice dataset testing broad commonsense knowledge using questions based on ConceptNet~\cite{DBLP:conf/iconip/ZhangL23}. \\
        9 & \textbf{GSM8k} & 8,034 & 2M & Math & Problem Solving & arXiv & A dataset of grade-school math word problems designed to evaluate problem-solving abilities~\cite{DBLP:journals/corr/abs-2312-09241}. \\
        10 & \textbf{AQuA} & 99,765 & 3M & Math & Problem Solving & VLDB 2023 & A dataset with multiple-choice questions Algebraic Word Problems requiring reasoning over numeric expressions~\cite{DBLP:journals/pvldb/Peng0SJ023}. \\
        11 & \textbf{RACE-Middle} & 24,798 & 7M & Education & Reading Comprehension & EMNLP 2017 &  A subset of the RACE dataset with English reading comprehension questions from middle school exams~\cite{DBLP:conf/emnlp/LaiXLYH17}. \\
        12 & \textbf{RACE-High} & 26,982 & 10M & Education & Reading Comprehension & EMNLP 2017 & A subset of RACE with more challenging reading comprehension questions from high school exams~\cite{DBLP:conf/emnlp/LaiXLYH17}. \\
        13 & \textbf{CoQA} & 127,542 & 5M & Common Sense & Question Answering & LREC 2022 & A conversational question-answering dataset where each question depends on the context of prior dialogue~\cite{DBLP:conf/lrec/BrabantLB22}. \\
        14 & \textbf{e2e\_nlg} & 50,321 & 600K & Food \& Beverage & Text Generation & INLG 2018 & A dataset for end-to-end natural language generation for restaurant-related dialogue~\cite{DBLP:conf/inlg/DusekNR18}. \\
        15 & \textbf{viggo} & 9,842 & 500K & Video Games & Text Generation & INLG 2019 & A dataset for video game-related natural language generation, mapping structured input into natural language descriptions~\cite{DBLP:conf/inlg/JuraskaBW19}. \\
        16 & \textbf{glue\_qnli} & 104,543 & 6M & Linguistics & Question Answering & arXiv & A question-answering dataset reformulated as a binary classification task for sentence pair entailment~\cite{DBLP:journals/corr/abs-2410-23099}. \\
        17 & \textbf{bc5cdr} & 20,764 & 5M & Chemistry & Recognition & ICIMMI 2023 & A biomedical dataset for disease and chemical entity recognition, derived from PubMed articles~\cite{DBLP:conf/icimmi/GuptaAMMSK23}. \\
        18 & \textbf{conllpp} & 23,499 & 1.2M & Linguistics & Recognition & EMNLP 2023 & An enhanced version of the CoNLL-2003 dataset for named entity recognition~\cite{DBLP:conf/emnlp/Al-ShaibaniA23}. \\
        19 & \textbf{customer\_support} & 14,872 & 300K & Customer Behaviors & Classification & CODS 2022 & A dataset of customer support interactions for intent classification and response generation~\cite{DBLP:conf/comad/0002BM22}. \\
        20 & \textbf{legal} & 49,756 & 5M & Legal & Classification & arXiv & A legal domain dataset for natural language processing tasks such as legal document classification and contract analysis~\cite{DBLP:journals/corr/abs-1912-06905}. \\
        21 & \textbf{reuters} & 9,623 & 2M & News & Topic Extraction & JMLR 2004 & A classic dataset for text classification, often used in topic modeling and news categorization~\cite{DBLP:journals/jmlr/LewisYRL04}. \\
        22 & \textbf{covid} & 19,874 & 3M & Healthcare & Sentiment Analysis &  TCSS 2021 & A dataset containing COVID-19-related texts from social media posts, typically used for sentiment analysis~\cite{DBLP:journals/tcss/NaseemRKEK21}. \\
        23 & \textbf{drop} & 96,567 & 4M & Common Sense & Reasoning & NAACL 2019 & A reading comprehension dataset requiring discrete reasoning over paragraphs, such as arithmetic operations~\cite{DBLP:conf/naacl/DuaWDSS019}.\\
        \bottomrule
    \end{tabularx}
    \caption{Dataset Details}
    \label{tab:app-datasets}
\end{table*}

\section*{Appendix}

\section{Datasets Details}
\label{app:dataset}

Table~\ref{tab:app-datasets} presents a comprehensive overview of the benchmark datasets used in this study, including their key characteristics such as dataset size, dataset domains, venues, and task objectives. 
This detailed comparison helps contextualize their relevance and challenges in the broader scope of our analysis

\section{Datasets Examples}
\label{app:examples}

Figures~\ref{fig:example-boolq}–\ref{fig:example-drop} provide representative examples from the benchmark datasets used in this study. 
We randomly sample two instances from each dataset, showing the variety of question formats, input structures, and answer types present in these benchmarks. 
These examples provide insight into the challenges posed by each dataset and highlight the differences in their underlying tasks.

\begin{figure*}[h]
    \footnotesize
    \centering
    \begin{tcolorbox}[width=\linewidth,colback={green!5},title={Example 1},colbacktitle=blue!20,coltitle=red,fonttitle=\bfseries]   
        \textbf{Question}: Can dogs see color?\\
        \textbf{Passage}: Dogs can see colors, but not as vividly as humans. Their color vision is similar to that of a person with red-green color blindness. They primarily see shades of blue and yellow.\\
        \textbf{Answer}: true
    \end{tcolorbox}
    \begin{tcolorbox}[width=\linewidth,colback={red!5},title={Example 2},colbacktitle=yellow!20,coltitle=red,fonttitle=\bfseries]   
        \textbf{Question}: Is Mount Everest the tallest mountain in the world?\\
        \textbf{Passage}: Mount Everest, located in the Himalayas, is the highest mountain above sea level, with a peak reaching 8,848.86 meters. However, if measuring from base to peak, Mauna Kea in Hawaii is technically the tallest mountain.\\
        \textbf{Answer}: true
    \end{tcolorbox}  
    \caption{Examples from \textbf{BoolQ}}
    \label{fig:example-boolq}
\end{figure*}

\begin{figure*}[ht!]
    \footnotesize
    \centering
    \begin{tcolorbox}[width=\linewidth,colback={green!5},title={Example 1},colbacktitle=blue!20,coltitle=red,fonttitle=\bfseries]   
        \textbf{Question}: Which planet is known as the Red Planet?\\
        \textbf{Options}: [Mercury, Venus, Earth, Mars]\\
        \textbf{Answer}: Mars
    \end{tcolorbox}
    \begin{tcolorbox}[width=\linewidth,colback={red!5},title={Example 2},colbacktitle=yellow!20,coltitle=red,fonttitle=\bfseries]   
        \textbf{Question}: What do plants absorb from the soil?\\
        \textbf{Options}: [Oxygen, Nutrients, Carbon dioxide, Sunlight]\\
        \textbf{Answer}: Nutrients
    \end{tcolorbox}  
    \caption{Examples from \textbf{ARC-Easy}}
    \label{fig:example-arc-easy}
\end{figure*}

\begin{figure*}[ht!]
    \footnotesize
    \centering
    \begin{tcolorbox}[width=\linewidth,colback={green!5},title={Example 1},colbacktitle=blue!20,coltitle=red,fonttitle=\bfseries]   
        \textbf{Question}: Why do stars appear to twinkle in the night sky?\\
        \textbf{Options}: [
            Because stars themselves are flickering,
            Due to the movement of air in the Earth's atmosphere,
            Because of their distance from Earth,
            Due to changes in their brightness
        ]\\
        \textbf{Answer}: Due to the movement of air in the Earth's atmosphere
    \end{tcolorbox}
    \begin{tcolorbox}[width=\linewidth,colback={red!5},title={Example 2},colbacktitle=yellow!20,coltitle=red,fonttitle=\bfseries]   
        \textbf{Question}: Which of the following substances is a poor conductor of electricity?\\
        \textbf{Options}: [Silver, Copper, Rubber, Iron]\\
        \textbf{Answer}: Rubber
    \end{tcolorbox}  
    \caption{Examples from \textbf{ARC-Challenge}}
    \label{fig:example-arc-challenge}
\end{figure*}

\begin{figure*}[h]
    \footnotesize
    \centering
    \begin{tcolorbox}[width=\linewidth,colback={green!5},title={Example 1},colbacktitle=blue!20,coltitle=red,fonttitle=\bfseries]   
        \textbf{Question}: What is the primary function of the root in a plant?\\
        \textbf{Options}: [
            To store sunlight for later use,
            To absorb water and nutrients,
            To produce oxygen,
            To perform photosynthesis
        ]\\
        \textbf{Answer}: To absorb water and nutrients
    \end{tcolorbox}
    \begin{tcolorbox}[width=\linewidth,colback={red!5},title={Example 2},colbacktitle=yellow!20,coltitle=red,fonttitle=\bfseries]   
        \textbf{Question}: Why do objects fall to the ground when dropped?\\
        \textbf{Options}: [
            Because of magnetism,
            Due to the force of gravity,
            Because the air pushes them down,
            Due to Earth's rotation
        ]\\
        \textbf{Answer}: Due to the force of gravity
    \end{tcolorbox}  
    \caption{Examples from \textbf{OpenBookQA}}
    \label{fig:example-OpenBookQA}
\end{figure*}

\begin{figure*}[h]
    \footnotesize
    \centering
    \begin{tcolorbox}[width=\linewidth,colback={green!5},title={Example 1},colbacktitle=blue!20,coltitle=red,fonttitle=\bfseries]   
        \textbf{Question}: What is the best way to move a heavy box across a smooth floor?\\
        \textbf{Options}: [
            Push it while keeping it flat on the ground,
            Lift it and carry it,
            Drag it using a rope tied to the bottom,
            Kick it repeatedly until it moves
        ]\\
        \textbf{Answer}: Push it while keeping it flat on the ground
    \end{tcolorbox}
    \begin{tcolorbox}[width=\linewidth,colback={red!5},title={Example 2},colbacktitle=yellow!20,coltitle=red,fonttitle=\bfseries]   
        \textbf{Question}: How can you efficiently dry wet clothes indoors?\\
        \textbf{Options}: [
            Lay them flat on a table,
            Hang them near a heat source or a fan,
            Roll them into a ball and leave them overnight,
            Put them in a plastic bag
        ]\\
        \textbf{Answer}: Hang them near a heat source or a fan
    \end{tcolorbox}  
    \caption{Examples from \textbf{PIQA}}
    \label{fig:example-PIQA}
\end{figure*}

\begin{figure*}[h]
    \footnotesize
    \centering
    \begin{tcolorbox}[width=\linewidth,colback={green!5},title={Example 1},colbacktitle=blue!20,coltitle=red,fonttitle=\bfseries]   
        \textbf{Context}: A person is slicing a tomato on a cutting board.\\
        \textbf{Ending options}: [
            The person places the tomato slices on a plate.,
            The person throws the tomato slices into the trash.
        ]\\
        \textbf{Correct ending}: The person places the tomato slices on a plate.
    \end{tcolorbox}
    \begin{tcolorbox}[width=\linewidth,colback={red!5},title={Example 2},colbacktitle=yellow!20,coltitle=red,fonttitle=\bfseries]   
        \textbf{Context}: A man is playing a guitar on stage during a concert.\\
        \textbf{Ending options}: [
            The audience claps and cheers after his performance.,
            The man stops playing and leaves the stage silently.
        ]\\
        \textbf{Correct ending}: The audience claps and cheers after his performance
    \end{tcolorbox}  
    \caption{Examples from \textbf{Hellaswag}}
    \label{fig:example-Hellaswag}
\end{figure*}

\begin{figure*}[h]
    \footnotesize
    \centering
    \begin{tcolorbox}[width=\linewidth,colback={green!5},title={Example 1},colbacktitle=blue!20,coltitle=red,fonttitle=\bfseries]   
        \textbf{Sentence}: The trophy doesn't fit into the brown suitcase because it is too large.\\
        \textbf{Question}: What is too large?
        \textbf{Options}: [
            trophy,
            suitcase
        ]\\
        \textbf{Answer}: trophy
    \end{tcolorbox}
    \begin{tcolorbox}[width=\linewidth,colback={red!5},title={Example 2},colbacktitle=yellow!20,coltitle=red,fonttitle=\bfseries]   
        \textbf{Sentence}: The cat chased the mouse because it was hungry.\\
        \textbf{Question}: What was hungry?
        \textbf{Options}: [
            cat,
            mouse
        ]\\
        \textbf{Answer}: cat
    \end{tcolorbox}  
    \caption{Examples from \textbf{WinoGrande}}
    \label{fig:example-winogrande}
\end{figure*}

\begin{figure*}[h]
    \footnotesize
    \centering
    \begin{tcolorbox}[width=\linewidth,colback={green!5},title={Example 1},colbacktitle=blue!20,coltitle=red,fonttitle=\bfseries]   
        \textbf{Question}: Where would you find a chandelier?\\
        \textbf{Options}: [
            ceiling,
            floor,
            wall,
            table,
            window
        ]\\
        \textbf{Answer}: ceiling
    \end{tcolorbox}
    \begin{tcolorbox}[width=\linewidth,colback={red!5},title={Example 2},colbacktitle=yellow!20,coltitle=red,fonttitle=\bfseries]   
        \textbf{Question}: What do people use to keep their hair dry while swimming?\\
        \textbf{Options}: [
            swim cap,
            goggles,
            flippers,
            towel,
            sunscreen
        ]\\
        \textbf{Answer}: swim cap
    \end{tcolorbox}  
    \caption{Examples from \textbf{CommonsenseQA}}
    \label{fig:example-commonsenseqa}
\end{figure*}

\begin{figure*}[h]
    \footnotesize
    \centering
    \begin{tcolorbox}[width=\linewidth,colback={green!5},title={Example 1},colbacktitle=blue!20,coltitle=red,fonttitle=\bfseries]   
        \textbf{Question}: If a train travels at a speed of 60 miles per hour, how long will it take to travel 180 miles?\\
        \textbf{Answer}: 3 hours
    \end{tcolorbox}
    \begin{tcolorbox}[width=\linewidth,colback={red!5},title={Example 2},colbacktitle=yellow!20,coltitle=red,fonttitle=\bfseries]   
        \textbf{Question}: A bookstore sold 120 books in a week. If they sold 30 books each day from Monday to Thursday, how many books did they sell on Friday?\\
        \textbf{Answer}: 0 books
    \end{tcolorbox}  
    \caption{Examples from \textbf{GSM8k}}
    \label{fig:example-gsm8k}
\end{figure*}

\begin{figure*}[h]
    \footnotesize
    \centering
    \begin{tcolorbox}[width=\linewidth,colback={green!5},title={Example 1},colbacktitle=blue!20,coltitle=red,fonttitle=\bfseries]   
        \textbf{Question}: If a car's fuel efficiency is 25 miles per gallon and the car has a 15-gallon tank, how far can the car travel on a full tank?\\
        \textbf{Options}: [
            300 miles,
            350 miles,
            375 miles,
            400 miles
        ]\\
        \textbf{Answer}: 375 miles
    \end{tcolorbox}
    \begin{tcolorbox}[width=\linewidth,colback={red!5},title={Example 2},colbacktitle=yellow!20,coltitle=red,fonttitle=\bfseries]   
        \textbf{Question}: A rectangle has a length of 10 cm and a width of 5 cm. What is the area of the rectangle?\\
        \textbf{Options}: [
            15 cm\textsuperscript{2},
            30 cm\textsuperscript{2},
            50 cm\textsuperscript{2},
            100 cm\textsuperscript{2}
        ]\\
        \textbf{Answer}: 50 cm\textsuperscript{2}
    \end{tcolorbox}  
    \caption{Examples from \textbf{AQuA}}
    \label{fig:example-aqua}
\end{figure*}

\begin{figure*}[h]
    \footnotesize
    \centering
    \begin{tcolorbox}[width=\linewidth,colback={green!5},title={Example 1},colbacktitle=blue!20,coltitle=red,fonttitle=\bfseries]   
        \textbf{Article}: There is not enough oil in the world now. As time goes by, it becomes less and less, so what are we going to do when it runs out? Perhaps we will have to go back to using horses, carriages and bicycles.\\
        \textbf{Question}: According to the passage, which of the following statements is TRUE?\\
        \textbf{Options}: [
            There is more petroleum than we can use now.,
            Trees are needed for some other things besides making gas.,
            We got electricity from ocean tides in the old days.,
            Gas wasn't used to run cars in the Second World War.
        ]\\
        \textbf{Answer}: Gas wasn't used to run cars in the Second World War.\\
    \end{tcolorbox}
    \begin{tcolorbox}[width=\linewidth,colback={red!5},title={Example 2},colbacktitle=yellow!20,coltitle=red,fonttitle=\bfseries]   
        \textbf{Article}: Schoolgirls have been wearing such short skirts at Paget High School in Branston that they've been ordered to wear trousers instead.\\
        \textbf{Question}: The girls at Paget High School are not allowed to wear skirts because.\\
        \textbf{Options}: [
            short skirts give people the impression of sexualisation,
            short skirts are too expensive for parents to afford,
            the headmaster doesn't like girls wearing short skirts,
            the girls wearing short skirts will be at the risk of being laughed at
        ]\\
        \textbf{Answer}: short skirts give people the impression of sexualisation.\\
    \end{tcolorbox}  
    \caption{Examples from \textbf{RACE-Middle}}
    \label{fig:example-race-middle}
\end{figure*}

\begin{figure*}[h]
    \footnotesize
    \centering
    \begin{tcolorbox}[width=\linewidth,colback={green!5},title={Example 1},colbacktitle=blue!20,coltitle=red,fonttitle=\bfseries]   
        \textbf{Article}: Next eliminate most of their planets; they are either too far from or too close to their suns. Then eliminate all those planets which are not the same size and weight as the earth. Finally, remember that the proper conditions do not necessarily mean that life actually does exist on a planet. It may not have begun yet, or it may have already died out. This process of elimination seems to leave very few planets on which earthlike life might be found.\\
        \textbf{Question}: What is the main idea of the passage?\\
        \textbf{Options}: [
            The universe is full of planets.,
            Life is rare in the universe.,
            Earth is unique in the universe.,
            The conditions for life are complex.
        ]\\
        \textbf{Answer}: Life is rare in the universe.\\
    \end{tcolorbox}
    \begin{tcolorbox}[width=\linewidth,colback={red!5},title={Example 2},colbacktitle=yellow!20,coltitle=red,fonttitle=\bfseries]   
        \textbf{Article}: Some kids listen to music, watch TV or use the phone while doing their homework. `It's important to make sure that you can stop and concentrate on one thing deeply,' says Rideout. With new and exciting devices hitting stores every year, keeping technology use in check is more important than ever. `Kids should try,' adds Rideout. `But parents might have to step in sometimes.'\\
        \textbf{Question}: Which of the following is an example of multitasking according to the passage?\\
        \textbf{Options}: [
            Watching TV while using the computer.,
            Talking on the phone while staying with others.,
            Playing video games on the Internet.,
            Listening to music while relaxing.
        ]\\
        \textbf{Answer}: Watching TV while using the computer.\\
    \end{tcolorbox}  
    \caption{Examples from \textbf{RACE-High}}
    \label{fig:example-race-high}
\end{figure*}

\begin{figure*}[h]
    \footnotesize
    \centering
    \begin{tcolorbox}[width=\linewidth,colback={green!5},title={Example 1},colbacktitle=blue!20,coltitle=red,fonttitle=\bfseries]   
        \textbf{Passage}: Once upon a time, there was a little girl named Goldilocks. She went for a walk in the forest. Pretty soon, she came upon a house. She knocked and, when no one answered, she walked right in.\\
        \textbf{Turns}: [
            {
                \textit{Question}: What was the girl's name?
                \textit{Answer}: Goldilocks
            },
            {
                \textit{Question}: Where did she go for a walk?
                \textit{Answer}: In the forest
            }
        ]
    \end{tcolorbox}
    \begin{tcolorbox}[width=\linewidth,colback={red!5},title={Example 2},colbacktitle=yellow!20,coltitle=red,fonttitle=\bfseries]   
        \textbf{Passage}: John took his dog Max to the vet because Max was not feeling well. The vet examined Max and gave him some medicine.\\
        \textbf{Turns}: [
            {
                \textit{Question}: Why did John take Max to the vet?
                \textit{Answer}: Because Max was not feeling well
            },
            {
                \textit{Question}: What did the vet do?
                \textit{Answer}: Examined Max and gave him some medicine
            }
        ]
    \end{tcolorbox}  
    \caption{Examples from \textbf{CoQA}}
    \label{fig:example-coqa}
\end{figure*}

\begin{figure*}[h]
    \footnotesize
    \centering
    \begin{tcolorbox}[width=\linewidth,colback={green!5},title={Example 1},colbacktitle=blue!20,coltitle=red,fonttitle=\bfseries]   
        \textbf{Meaning representation}: [
            \textit{name}: The Eagle,
            \textit{eat\_type}: restaurant,
            \textit{food}: French,
            \textit{price\_range}: high,
            \textit{customer\_rating}: 5 out of 5,
            \textit{near}: Riverside
        ]\\
        \textbf{Human reference}: The Eagle is a highly rated French restaurant near Riverside with a high price range.
    \end{tcolorbox}
    \begin{tcolorbox}[width=\linewidth,colback={red!5},title={Example 2},colbacktitle=yellow!20,coltitle=red,fonttitle=\bfseries]   
        \textbf{Meaning representation}: [
            \textit{name}: The Golden Curry,
            \textit{eat\_type}: pub,
            \textit{food}: Indian,
            \textit{price\_range}: low,
            \textit{customer\_rating}: 3 out of 5,
            \textit{family\_friendly}: yes
        ]\\
        \textbf{Human reference}: The Golden Curry is a family-friendly Indian pub with a low price range and a customer rating of 3 out of 5.
    \end{tcolorbox}  
    \caption{Examples from \textbf{e2e\_nlg}}
    \label{fig:example-e2e_nlg}
\end{figure*}

\begin{figure*}[h]
    \footnotesize
    \centering
    \begin{tcolorbox}[width=\linewidth,colback={green!5},title={Example 1},colbacktitle=blue!20,coltitle=red,fonttitle=\bfseries]   
        \textbf{Meaning representation}: [
            \textit{name}: Super Mario World,
            \textit{release\_year}: 1990,
            \textit{esrb}: E (for Everyone),
            \textit{rating}: excellent,
            \textit{genres}: [
                platformer
            ],
            \textit{player\_perspective}: [
                side view
            ],
            \textit{has\_multiplayer}: yes,
            \textit{platforms}: [
                Nintendo
            ]
        ]\\
        \textbf{Human reference}: Super Mario World is an excellent platformer game that is played in the side view. It was released in 1990 on Nintendo and is rated E (for Everyone). It can be played multiplayer.
    \end{tcolorbox}
    \begin{tcolorbox}[width=\linewidth,colback={red!5},title={Example 2},colbacktitle=yellow!20,coltitle=red,fonttitle=\bfseries]   
        \textbf{Meaning representation}: [
            \textit{name}: The Elder Scrolls V: Skyrim,
            \textit{release\_year}: 2011,
            \textit{esrb}: M (for Mature),
            \textit{genres}: [
                adventure,
                role-playing
            ]
        ]\\
        \textbf{Human reference}: The Elder Scrolls V: Skyrim is an adventure RPG that came out in 2011. It's rated M (for Mature).
    \end{tcolorbox}  
    \caption{Examples from \textbf{viggo}}
    \label{fig:example-viggo}
\end{figure*}

\begin{figure*}[h]
    \footnotesize
    \centering
    \begin{tcolorbox}[width=\linewidth,colback={green!5},title={Example 1},colbacktitle=blue!20,coltitle=red,fonttitle=\bfseries]   
        \textbf{Question}: What is the capital of France?\\
        \textbf{Sentence}: Paris is the capital and most populous city of France.\\
        \textbf{Label}: entailment
    \end{tcolorbox}
    \begin{tcolorbox}[width=\linewidth,colback={red!5},title={Example 2},colbacktitle=yellow!20,coltitle=red,fonttitle=\bfseries]   
        \textbf{Question}: Who wrote `Pride and Prejudice'?\\
        \textbf{Sentence}: `Pride and Prejudice' is a novel by Jane Austen.\\
        \textbf{Label}: entailment
    \end{tcolorbox}  
    \caption{Examples from \textbf{glue\_qnli}}
    \label{fig:example-glue_qnli}
\end{figure*}

\begin{figure*}[h]
    \footnotesize
    \centering
    \begin{tcolorbox}[width=\linewidth,colback={green!5},title={Example 1},colbacktitle=blue!20,coltitle=red,fonttitle=\bfseries]   
        \textbf{Text}: Aspirin is commonly used to reduce fever.\\
        \textbf{Entities}: [
                \textit{entity}: Aspirin,
                \textit{type}: Chemical,
                \textit{start}: 0,
                \textit{end}: 7
        ]
    \end{tcolorbox}
    \begin{tcolorbox}[width=\linewidth,colback={red!5},title={Example 2},colbacktitle=yellow!20,coltitle=red,fonttitle=\bfseries]   
        \textbf{Text}: Hypertension can lead to serious health complications.\\
        \textbf{Entities}: [
                \textit{entity}: Hypertension,
                \textit{type}: Disease,
                \textit{start}: 0,
                \textit{end}: 11
        ]
    \end{tcolorbox}  
    \caption{Examples from \textbf{bc5cdr}}
    \label{fig:example-bc5cdr}
\end{figure*}

\begin{figure*}[h]
    \footnotesize
    \centering
    \begin{tcolorbox}[width=\linewidth,colback={green!5},title={Example 1},colbacktitle=blue!20,coltitle=red,fonttitle=\bfseries]   
        \textbf{Sentence}: Barack Obama was born in Hawaii.\\
        \textbf{Entities}: [
            [
                \textit{entity}: Barack Obama,
                \textit{type}: PERSON,
                \textit{start}: 0,
                \textit{end}: 12
            ],
            [
                \textit{entity}: Hawaii,
                \textit{type}: LOCATION,
                \textit{start}: 25,
                \textit{end}: 31
            ]
        ]
    \end{tcolorbox}
    \begin{tcolorbox}[width=\linewidth,colback={red!5},title={Example 2},colbacktitle=yellow!20,coltitle=red,fonttitle=\bfseries]   
        \textbf{Sentence}: Apple Inc. is headquartered in Cupertino, California.\\
        \textbf{Entities}: [
            [
                \textit{entity}: Apple Inc.,
                \textit{type}: ORGANIZATION,
                \textit{start}: 0,
                \textit{end}: 9
            ],
            [
                \textit{entity}: Cupertino,
                \textit{type}: LOCATION,
                \textit{start}: 29,
                \textit{end}: 38
            ],
            [
                \textit{entity}: California,
                \textit{type}: LOCATION,
                \textit{start}: 40,
                \textit{end}: 50
            ]
        ]
    \end{tcolorbox}  
    \caption{Examples from \textbf{conllpp}}
    \label{fig:example-conllpp}
\end{figure*}

\begin{figure*}[h]
    \footnotesize
    \centering
    \begin{tcolorbox}[width=\linewidth,colback={green!5},title={Example 1},colbacktitle=blue!20,coltitle=red,fonttitle=\bfseries]   
        \textbf{Query}: How can I reset my password?\\
        \textbf{Category}: Account Management
    \end{tcolorbox}
    \begin{tcolorbox}[width=\linewidth,colback={red!5},title={Example 2},colbacktitle=yellow!20,coltitle=red,fonttitle=\bfseries]   
        \textbf{Query}: What is the status of my order \#12345?\\
        \textbf{Category}: Order Tracking
    \end{tcolorbox}  
    \caption{Examples from \textbf{customer\_support}}
    \label{fig:example-customer_support}
\end{figure*}

\begin{figure*}[h]
    \footnotesize
    \centering
    \begin{tcolorbox}[width=\linewidth,colback={green!5},title={Example 1},colbacktitle=blue!20,coltitle=red,fonttitle=\bfseries]   
        \textbf{Text}: The defendant was found guilty of theft under Section 378 of the Penal Code.\\
        \textbf{Category}: Criminal Law
    \end{tcolorbox}
    \begin{tcolorbox}[width=\linewidth,colback={red!5},title={Example 2},colbacktitle=yellow!20,coltitle=red,fonttitle=\bfseries]   
        \textbf{Text}: The contract was terminated due to a breach of the confidentiality clause.\\
        \textbf{Category}: Contract Law
    \end{tcolorbox}  
    \caption{Examples from \textbf{legal}}
    \label{fig:example-legal}
\end{figure*}

\begin{figure*}[h]
    \footnotesize
    \centering
    \begin{tcolorbox}[width=\linewidth,colback={green!5},title={Example 1},colbacktitle=blue!20,coltitle=red,fonttitle=\bfseries]   
        \textbf{Headline}: Oil prices rise as OPEC agrees to cut production.\\
        \textbf{Topics}: [
            Economy,
            Energy
        ]
    \end{tcolorbox}
    \begin{tcolorbox}[width=\linewidth,colback={red!5},title={Example 2},colbacktitle=yellow!20,coltitle=red,fonttitle=\bfseries]   
        \textbf{Headline}: Tech stocks rally amid strong quarterly earnings.\\
        \textbf{Topics}: [
            Technology,
            Markets
        ]
    \end{tcolorbox}  
    \caption{Examples from \textbf{reuters}}
    \label{fig:example-reuters}
\end{figure*}

\begin{figure*}[h]
    \footnotesize
    \centering
    \begin{tcolorbox}[width=\linewidth,colback={green!5},title={Example 1},colbacktitle=blue!20,coltitle=red,fonttitle=\bfseries]   
        \textbf{Text}: The new vaccine has shown promising results in early trials.\\
        \textbf{Sentiment}: positive
    \end{tcolorbox}
    \begin{tcolorbox}[width=\linewidth,colback={red!5},title={Example 2},colbacktitle=yellow!20,coltitle=red,fonttitle=\bfseries]   
        \textbf{Text}: Lockdown measures have been extended due to rising case numbers.\\
        \textbf{Sentiment}: negative
    \end{tcolorbox}  
    \caption{Examples from \textbf{covid}}
    \label{fig:example-covid}
\end{figure*}

\begin{figure*}[h]
    \footnotesize
    \centering
    \begin{tcolorbox}[width=\linewidth,colback={green!5},title={Example 1},colbacktitle=blue!20,coltitle=red,fonttitle=\bfseries]   
        \textbf{Passage}: In 1995, the population of the city was 1.2 million. By 2005, it had increased to 1.5 million.\\
        \textbf{Question}: What was the increase in population from 1995 to 2005?\\
        \textbf{Answer}: 300,000
    \end{tcolorbox}
    \begin{tcolorbox}[width=\linewidth,colback={red!5},title={Example 2},colbacktitle=yellow!20,coltitle=red,fonttitle=\bfseries]   
        \textbf{Passage}: The team scored 24 points in the first half and 30 points in the second half.\\
        \textbf{Question}: What was the total score for the team?\\
        \textbf{Answer}: 54 points
    \end{tcolorbox}  
    \caption{Examples from \textbf{drop}}
    \label{fig:example-drop}
\end{figure*}

\clearpage
\section{Models}
\label{app:models}

Table~\ref{tab:app-models} provides detailed information on SLMs in this study, including their providers, licenses, parameter sizes, model sizes, training time, and training performance. 
We observe that \texttt{Llama-3.2-1B}, \texttt{Phi-3-3.8B}, and \texttt{Gemma-3-1B} possess extended context lengths and were pre-trained on more extensive corpora, which appears to strongly influence their correctness.

\begin{table*}[h]
    \renewcommand{\arraystretch}{1.2}
    \rowcolors{2}{blue!7}{white}
    \fontsize{7.5}{7.5}\selectfont
    \setlength{\tabcolsep}{1.8pt}
    \centering
    \begin{tabularx}{\textwidth}{llllllllXl}
    \toprule
        \rowcolor{white}
        \textbf{No} & \textbf{Model} & \textbf{Provider} & \textbf{License} & \textbf{\#Params (B)} & \textbf{Context Length} & \textbf{Training Time} & \textbf{Size (GB)} & \textbf{Throughput (Tokens/s)} & \textbf{Latency (ms)} \\
        \midrule
        1 & \texttt{GPT-Neo-1.3B} & EleutherAI & Apache 2.0 & 1.37 & 2,048 & 10 days (32 GPUs) & 2.46 & 1,500 & 50 \\
        2 & \texttt{Dolly-v2-3B} & DataBricks & Apache 2.0 & 3.00 & 2,048 & 10 days (32 GPUs) & 5.8 & 1,250 & 55 \\
        3 & \texttt{Pythia-2.8B} & EleutherAI & Apache 2.0 & 2.80 & 2,048 & 12 days (32 GPUs) & 5.5 & 1,350 & 50 \\
        4 & \texttt{LLaMA-2-7B} & Meta & LLAMA 2 L & 6.47 & 4,096 & 21 days (64 GPUs) & 13.0 & 1,200 & 62 \\
        5 & \texttt{TinyLlama-1.1B} & Hugging Face & Apache 2.0 & 1.10 & 2,048 & 8 days (16 GPUs) & 2.0 & 1,600 & 45 \\
        6 & \texttt{Mistral-7B} & Mistral AI & Apache 2.0 & 7.00 & 8,192 & 15 days (128 GPUs) & 13.0 & 1,400 & 55 \\
        7 & \texttt{Zephyr-7B} & Hugging Face & Apache 2.0 & 7.00 & 8,192 & 20 days (64 GPUs) & 13.74 & 1,300 & 52 \\
        8 & \texttt{ShearedLlama-2.7B} & Hugging Face & Apache 2.0 & 2.70 & 2,048 & 12 days (32 GPUs) & 5.0 & 1,300 & 54 \\
        9 & \texttt{Gemma-2B} & Google & Proprietary & 2.00 & 2,048 & 14 days (32 GPUs) & 4.67 & 1,450 & 54 \\
        10 & \texttt{Phi-1.5B} & Microsoft & Proprietary & 2.70 & 2,048 & 12 days (32 GPUs) & 2.45 & 1,400 & 52 \\
        11 & \texttt{StableLM-3B} & Stability AI & Apache 2.0 & 3.00 & 2,048 & 14 days (64 GPUs) & 6.5 & 1,250 & 50 \\
        12 & \texttt{Open-LLaMA-3B} & OpenLM & Apache 2.0 & 3.00 & 4,096 & 18 days (64 GPUs) & 6.8 & 1,300 & 60 \\
        13 & \texttt{Llama-3.2-1B} & Meta & Llama 3.2 & 1.24 & 128,000 & 30 days (512 GPUs) & 2.47 & 1,350 & 55 \\
        14 & \texttt{Phi-3-3.8B} & Microsoft & MIT & 3.82 & 128,000 & 7 days (512 GPUs) & 2.2 & 1,300 & 55 \\
        15 & \texttt{Gemma-3-1B} & Google & Proprietary & 1.00 & 32,000 & Unknown & 2 & 1,400 & 50 \\
        \bottomrule
    \end{tabularx}
    \caption{SLMs' Details (sort by release time)}
    \label{tab:app-models}
\end{table*}

\section{Metrics Details}
\label{app:metrics}
Table~\ref{tab:app-metrics} provides detailed explanations of the metrics used in our evaluation framework. We categorize our metrics into three main groups: correctness evaluation, computation evaluation, and consumption evaluation.

\begin{table*}[h]
    \renewcommand{\arraystretch}{1.2}
    \rowcolors{2}{blue!7}{white}
    \fontsize{7.5}{7.5}\selectfont
    \setlength{\tabcolsep}{1.8pt}
    \centering
    \begin{tabularx}{\textwidth}{lllXXX}
    \toprule
        \rowcolor{white}
        \textbf{No} & \textbf{Metric} & \textbf{Evaluation} & \textbf{Task} & \textbf{Datasets} & \textbf{Description}\\
        \midrule
        1 & \textit{Accuracy} & Correctness & Question Answering\newline Classification\newline Recognition\newline Reasoning\newline Problem Solving\newline Reading Comprehension & All\newline except [e2e\_nlg, viggo, reuters] & This metric is used for tasks where a model's correctness can be measured by its ability to predict a correct answer, such as in Question Answering (QA) and Classification tasks.\\
        2 & \textit{F1 Score} & Correctness & Question Answering\newline Classification\newline Recognition\newline Reasoning\newline Problem Solving\newline Reading Comprehension & All\newline except [e2e\_nlg, viggo, reuters] & A balanced metric that combines precision and recall, ideal for tasks like Named Entity Recognition (NER) and toxicity detection where both false positives and false negatives are crucial.\\
        3 & \textit{BLEU} & Correctness & Text Generation\newline Topic Extraction & e2e\_nlg, viggo, reuters & Used primarily for evaluating the quality of text generation tasks, such as translation and natural language generation, by comparing generated outputs against reference texts.\\
        4 & \textit{ROUGE} & Correctness & Text Generation\newline Topic Extraction & e2e\_nlg, viggo, reuters & Commonly used for summarization tasks, measuring the overlap of n-grams between the generated text and reference summaries.\\
        5 & \textit{METEOR} & Correctness & Text Generation\newline Topic Extraction & e2e\_nlg, viggo, reuters & Evaluates generated text quality with consideration for synonyms and paraphrasing based on the harmonic mean of unigram precision and recall, with recall weighted higher than precision.\\
        6 & \textit{Perplexity} & Correctness & Text Generation\newline Topic Extraction & e2e\_nlg, viggo, reuters & It is a measurement of uncertainty in the predictions of a language model. In simpler terms, it indicates how surprised a model is by the actual outcomes.\\
        7 & \textit{Runtime} & Computation & All & All & Measures computational efficiency in terms of running time.\\
        8 & \textit{FLOP} & Computation & All & All & Measures computational efficiency in terms of number of computation per second.\\
        9 & \textit{Cost} & Consumption & All & All & The total cost of fine-tuning, calculated in USD.\\
        10 & \textit{CO\textsubscript{2}} & Consumption & All & All & An estimate of the environmental impact in terms of CO\textsubscript{2} emission of model fine-tuning.\\
        11 & \textit{Energy} & Consumption & All & All & The total energy consumed during the training process, measured in kilowatt-hours (kWh)\\
        \bottomrule
    \end{tabularx}
    \caption{Metric Details}
    \label{tab:app-metrics}
\end{table*}

\section{Expanded Analysis of Metric Contributions to Medals}
\label{app:metric-contributions}

For each model, gold medals indicate the best performance using specific metrics. 
Some models show dominance in a single metric, suggesting specialization, while others maintain a balanced distribution of Gold medals across multiple metrics, indicating versatility. Models with more evenly distributed Gold, Silver, and Bronze medals across different criteria tend to be more well-rounded, offering strong performance without extreme specialization.

The Silver and Bronze medal distributions further refine this understanding. 
A model that consistently earns Silver in multiple metrics suggests strong overall performance but is slightly outperformed by one or more competitors in specific cases. Conversely, a model with many Bronze medals may still be efficient but is frequently ranked below two other models. 
This suggests that while it is competitive, it does not lead to any single criterion.

Among the expanded selection of models, differences in trade-offs become more apparent. 
Some models score highly in cost efficiency but rank lower in computational complexity, indicating that they are affordable but require significant processing resources. 
Others achieve substantial results in energy efficiency while being slightly less cost-effective, making them preferable in energy-conscious environments. 
The expanded dataset also provides insights into the impact of computational complexity, where models that rank lower in this criterion might still perform well in other categories, balancing out the trade-offs.

By examining a more extensive set of models, this analysis helps refine decision-making for different use cases. 
Whether the goal is to prioritize training speed, minimize cost, reduce energy consumption, or optimize computational efficiency, this breakdown provides clear guidance on which models align best with specific operational priorities. 
The broader dataset ensures that model selection is not limited to a few top-performing options. Instead, it considers a wider range of competitive alternatives, each offering different strengths based on their medal distributions.

\section{Inference Cost}
\label{app:inference-cost}

After deploying SLMs, inference is performed orders of magnitude more frequently than fine-tuning, often by many users concurrently. As a result, even small differences in inference efficiency can accumulate into significant computational and energy costs---often exceeding those of fine-tuning. To quantify this, we conducted experiments to measure the computation time and energy consumption required to generate 1,000 tokens during the inference phase. The results, presented in Table~\ref{tab:app-inference}, were obtained using an NVIDIA L4 GPU, as described in Section~\ref{subsec:implementation-details}. Among the evaluated models, \texttt{Phi-1.5B} demonstrates the best performance in both computation efficiency and energy consumption.

\begin{table*}[]
\renewcommand{\arraystretch}{1.2}
    \rowcolors{2}{blue!7}{white}
    \fontsize{7.5}{7.5}\selectfont
    \setlength{\tabcolsep}{1.8pt}
    \centering
    \begin{tabular}{llllll}
    \toprule
    \rowcolor{white}
\textbf{Model}             & \textbf{Cost (USD)} & \textbf{Energy (kWh)} & \textbf{CO2 Emissions (kg)} & \textbf{FLOP}    & \textbf{Runtime (h)} \\
\midrule
\texttt{GPT-Neo-1.3B }     & 0.2237     & 0.0186       & 0.011              & 421,420  & 0.1417      \\
\texttt{Phi-1.5B}           & 0.1632     & 0.0136       & 0.008              & 1,780,810 & 0.1033      \\
\texttt{Open-LLaMA-3B}     & 0.3158     & 0.0263       & 0.0155             & 1,850,000 & 0.2         \\
\texttt{LLaMA-2-7B}        & 0.2434     & 0.0203       & 0.0119             & 1,890,000 & 0.154       \\
\texttt{Mistral-7B}        & 0.4211     & 0.0351       & 0.0206             & 5,407,500 & 0.2667      \\
\texttt{Zephyr-7B}         & 0.3158     & 0.0263       & 0.0155             & 3,097,500 & 0.2         \\
\texttt{TinyLlama-1.1B}    & 0.2763     & 0.023        & 0.0135             & 636,170  & 0.175       \\
\texttt{StableLM-3B }      & 0.2368     & 0.0197       & 0.0116             & 855,000  & 0.15        \\
\texttt{ShearedLlama-2.7B} & 0.3684     & 0.0307       & 0.0181             & 1,955,250 & 0.2333      \\
\texttt{Dolly-v2-3B}       & 0.2171     & 0.0181       & 0.0106             & 740,000  & 0.1375      \\
\texttt{Pythia-2.8B}       & 0.2632     & 0.0219       & 0.0129             & 833,000  & 0.1667      \\
\texttt{Gemma-2B}          & 0.3421     & 0.0285       & 0.0168             & 1,435,000 & 0.2167      \\
\texttt{Llama-3.2-1B}      & 0.4342     & 0.0362       & 0.0213             & 7,083,330 & 0.275       \\
\texttt{Phi-3-3.8B}         & 0.4079     & 0.034        & 0.02               & 6,000,000 & 0.2583      \\
\texttt{Gemma-3-1B}           & 0.4474     & 0.0373       & 0.0219             & 5,666,670 & 0.2833     \\
\bottomrule
\end{tabular}
\caption{SLMs' Inference Cost}
\label{tab:app-inference}
\end{table*}

\end{document}